\documentclass{article}

\usepackage{arxiv}

\usepackage[utf8]{inputenc} 
\usepackage[T1]{fontenc}    
\usepackage{hyperref}       
\usepackage{url}            
\usepackage{booktabs}       
\usepackage{amsfonts}       
\usepackage{nicefrac}       
\usepackage{microtype}      
\usepackage{lipsum}		
\usepackage{graphicx}
\usepackage[square,sort,comma,numbers]{natbib}
\usepackage{doi}
\usepackage{xcolor}
\usepackage{array}
\usepackage{amsmath}
\usepackage{enumitem}
\usepackage{float}
\usepackage{graphicx, subfigure}
\usepackage{subfigure}

\newcommand{\V}[1]{{\mathbf{#1}}} 

\usepackage{color}

\newcommand{\better}[1]{\underline{#1}}
\newcommand{\best}[1]{\textbf{#1}}

\title{Taming the Long Tail of Deep Probabilistic Forecasting}

\author{Jedrzej Kozerawski\footnotemark[1]\\
	Department of Computer Science and Engineering\\
	University of California, San Diego\\
	San Diego, CA \\
	\texttt{jkozerawski@ucsd.edu} \\
	\And
	Mayank Sharan\footnotemark[1]\\
	Department of Computer Science and Engineering\\
	University of California, San Diego\\
	San Diego, CA \\
	\texttt{msharan@ucsd.edu} \\
	\And
	Rose Yu \\
	Department of Computer Science and Engineering\\
	University of California, San Diego\\
	San Diego, CA \\
	\texttt{roseyu@ucsd.edu} \\
}

\date{}

\begin{document}
\maketitle

\footnotetext[1]{Equal contribution}

\begin{abstract}
	Deep probabilistic forecasting is gaining attention in numerous applications ranging from weather prognosis, through electricity consumption estimation, to autonomous vehicle trajectory prediction. However, existing approaches focus on improvements on the most common scenarios without addressing the performance on rare and difficult cases.  In this work, we identify a long tail behavior in the performance of state-of-the-art deep learning methods on probabilistic forecasting. We present two moment-based tailedness measurement concepts to improve performance on the difficult tail examples: Pareto Loss and Kurtosis Loss. Kurtosis loss is a symmetric measurement as the fourth moment about the mean of the loss distribution. Pareto loss is asymmetric measuring right tailedness, modeling the loss using a generalized Pareto distribution (GPD).  We demonstrate the performance of our approach on several real-world datasets including time series and spatiotemporal trajectories, achieving significant improvements on the tail examples.
\end{abstract}


\section{Introduction}
Forecasting is one of the most fundamental problems in time series and spatiotemporal data analysis, with broad applications in energy,  finance, and transportation. Deep learning models  \cite{li2019enhancing,   salinas2020deepar, rasul2021autoregressive}  have emerged as state-of-the-art approaches  for forecasting rich time series and spatiotemporal data. In several forecasting competitions such as M5  forecasting competition \cite{makridakis2020m5},  Argoverse motion forecasting challenge \cite{chang2019argoverse}, and IARAI Traffic4cast contest \cite{kreil2020surprising}, almost all the winning solutions are based on deep neural networks.  

Despite  the encouraging progress, we discover that the forecasting performance of deep learning models has \textit{long-tail behavior}. That means, a significant amount of samples are  very difficult to forecast.   Existing works often measure the forecasting performance by averaging across test samples. However, such an average performance measured by root mean square error (RMSE) or mean absolute error (MAE) can be misleading. A low RMSE or MAE may indicate good averaged performance, but it does not prevent the model from behaving disastrously in difficult scenarios.

\begin{figure}[t!]
\vskip 0.1in
\begin{center}
\includegraphics[width=0.6\columnwidth]{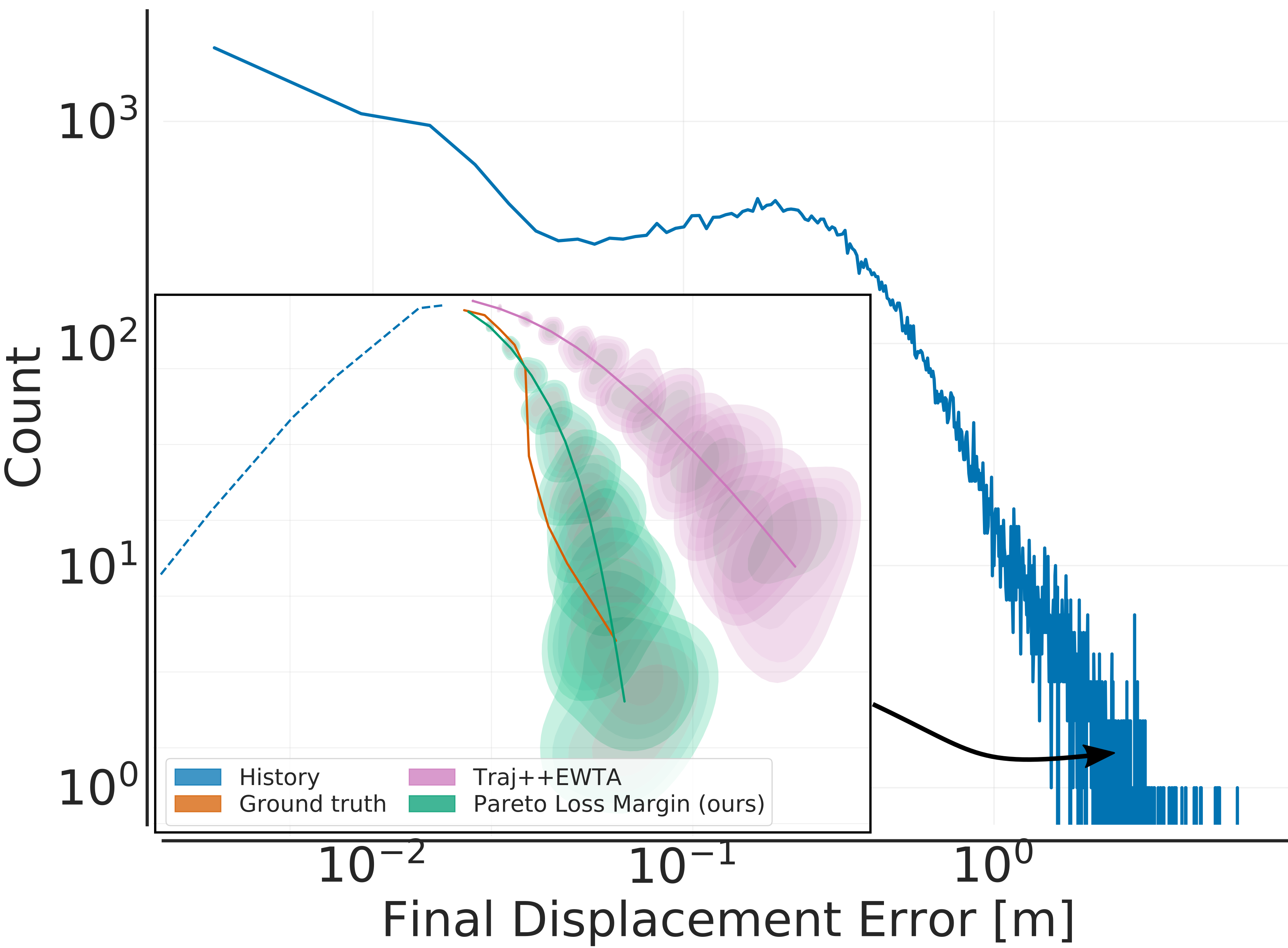}
\vskip -0.1in
\caption{Log-log plot of error distribution for trajectory prediction on the ETH-UCY dataset using SoTA(Traj++EWTA).
Also shown is a tail scenario with predictions using Traj++EWTA [purple] and Traj++EWTA+Pareto Margin Loss (ours) [teal].}
\label{fig:long-tail}
\end{center}
\vskip -0.1in
\end{figure}

From a practical perspective, the long-tail behavior in forecasting performance can be alarming. Figure \ref{fig:long-tail} visualizes examples of long-tail behavior for a motion forecasting task. In  motion forecasting,  the long-tail would correspond to  rare events in driving such as turning  maneuver and sudden stops. Failure to forecast accurately in these scenarios would pose paramount  safety risks in route planning. In electricity forecasting, the tail behavior would occur during short circuits, power outage, grid failures, or sudden behavior changes. Merely focusing on the average performance would ignore the electric load anomalies, significantly increasing the maintenance and operational cost.

Long-tailed learning is an area heavily studied in classification settings focusing on class imbalance. We refer readers to  Table 2 in \cite{menon2020long} and the survey paper by \cite{zhang2021deep} for a complete review. Most common approaches to address the long-tail problem include post-hoc normalization, data resampling, loss engineering, and learning class-agnostic representations. However, long-tail learning methods in classification are not directly translatable to forecasting as we do not have a pre-defined class. A recent work by \cite{makansi2021exposing} propose to use Kalman filter to gauge the difficulty of different forecasting examples  but such difficulties may not directly relate to deep neural networks used for the actual forecasting task.

In this paper, we address the long-tail behavior of prediction error for deep probabilistic forecasting.  We present two moment-based loss modifications: Kurtosis loss and Pareto loss.  Kurtosis is a well studied symmetric measure of tailedness as a scaled fourth moment of the distribution. Pareto loss uses Generalized Pareto Distribution (GPD) to fit the long-tailed error distribution and can also be described as a weighted summation of shifted moments.  We investigate these  tailedness measurements as regularization and loss weighting approaches for probabilistic forecasting tasks. We demonstrate significantly improved tail performance compared to the base model and the baselines, while achieving better average performance in most settings.

In summary, our contributions include
\begin{itemize}
    \item We discover long-tail behavior in the forecasting performance of deep probabilistic models.
    \item We investigate principled approaches to address long-tail behavior and propose two novel methods: Pareto loss and Kurtosis loss.
    \item We significantly improve the tail performance on four forecasting tasks including two time series and two spatiotemporal trajectory forecasting datasets.
\end{itemize}

\section{Related work}
\textbf{Deep probabilistic forecasting.} 
There is a flurry of work on using deep neural networks for probabilistic forecasting. For time series forecasting, a common practice is to combine classic time series models with deep learning, resulting in DeepAR \cite{salinas2020deepar}, Deep State Space \cite{rangapuram2018deep}, Deep Factors \cite{wang2019deep} and normalizing Kalman Filter \cite{de2020normalizing}. Others introduce normalizing flow \cite{rasul2020multivariate}, denoising diffusion \cite{rasul2021autoregressive} and particle filter \cite{pal2021rnn} to deep learning.  For trajectory forecasting, the majority of works focus on deterministic prediction. A few recent works propose to approximate the conditional distribution of future trajectories given the past with explicit parameterization \cite{mfp, luo2020probabilistic}, CVAE \cite{CVAE, desire, trajectron++} or implicit models such as GAN \cite{socialgan,liu2019naomi}. Nevertheless, most existing works focus on average performance, the issue of long-tail is largely overlooked in the community.

\textbf{Long-tailed learning.}
The main efforts for addressing the long-tail issue in learning revolve around reweighing, resampling, loss function engineering, and two-stage training, but mostly for classification.  Rebalancing during training comes either in form of synthetic minority oversampling~\cite{chawla2002smote}, oversampling with adversarial examples~\cite{Kozerawski_2020_ACCV}, inverse class frequency balancing~\cite{liu2019large}, balancing using effective number of samples~\cite{cui2019class}, or balance-oriented mixup augmentation~\cite{xu2021towards}. Another direction involves post-processing either in form of normalized calibration~\cite{pan2021model} or logit adjustment~\cite{menon2020long}. An important direction is loss modification approaches such as Focal Loss~\cite{lin2017focal}, Shrinkage Loss~\cite{lu2018deep}, and Balanced Meta-Softmax~\cite{ren2020balanced}. Others utilize two-stage training~\cite{liu2019large, cao2019learning} or separate expert networks~\cite{zhou2020bbn, li2020overcoming, wang2020long}. We refer the readers to~\cite{zhang2021deep} for an extensive survey. \cite{tang2020long} indicated SGD momentum can  contribute to the aggravation of the long-tail problem and suggested de-confounded training to mitigate its effects. \cite{feldman2020does, feldman2020neural} performed theoretical analysis and suggested label memorization in long-tail distribution as a necessity for the network to generalize.

A few were developed for imbalanced regression. Many approaches revolve around modifications of SMOTE such as adapted to regression SMOTER~\cite{torgo2013smote}, augmented with Gaussian Noise SMOGN~\cite{branco2017smogn}, or~\cite{ribeiro2020imbalanced} work extending for prediction of extremely rare values. \cite{steininger2021density} proposed DenseWeight, a method based on Kernel Density Estimation for better assessment of the relevance function for sample reweighing. \cite{yang2021delving} proposed a distribution smoothing over label (LDS) and feature space (FDS) for imbalanced regression. A concurrent work is \cite{makansi2021exposing} where they noticed the long-tail error distribution for trajectory prediction. They used Kalman filter~\cite{kalman1960new} performance as a difficulty measure and utilized contrastive learning to alleviate the tail problem. However, the tail of Kalman Filter may differ from that of deep learning models, which we elaborate on in later sections.

\section{Methodology}

We first identify the long-tail phenomena in probabilistic forecasting. Then, we propose two related strategies based on Pareto loss and Kurtosis loss to mitigate the tail issue.

\subsection{Long-tail in probabilistic forecasting}
\label{sec:method_long_tail}

Given input  $x_t\in\mathbb{R}^{d_{in}}$ and output $y_t\in\mathbb{R}^{d_{out}}$ respectively,  probabilistic forecasting task aims to predict the conditional distribution of future states $\V{y}=(y_{t+1}, \dotsc, y_{t+h})$ given current and past observations $\V{x}=(x_{t-k},\dotsc,x_t)$ as:
\begin{align}
p(y_{t+1}, \dotsc, y_{t+h}| x_{t-k},\dotsc,,x_t)
\end{align}
where $k$ is the length of the history and $h$ is the prediction horizon. We denote the maximum likelihood probabilistic forecasting model prediction as $\V{\hat{y}}=(\hat{y}_{t+1}, \dotsc, \hat{y}_{t+h})$. 

\begin{figure}[t!]
\vskip 0.1in
\begin{center}
\centerline{
\includegraphics[width=0.5\columnwidth]{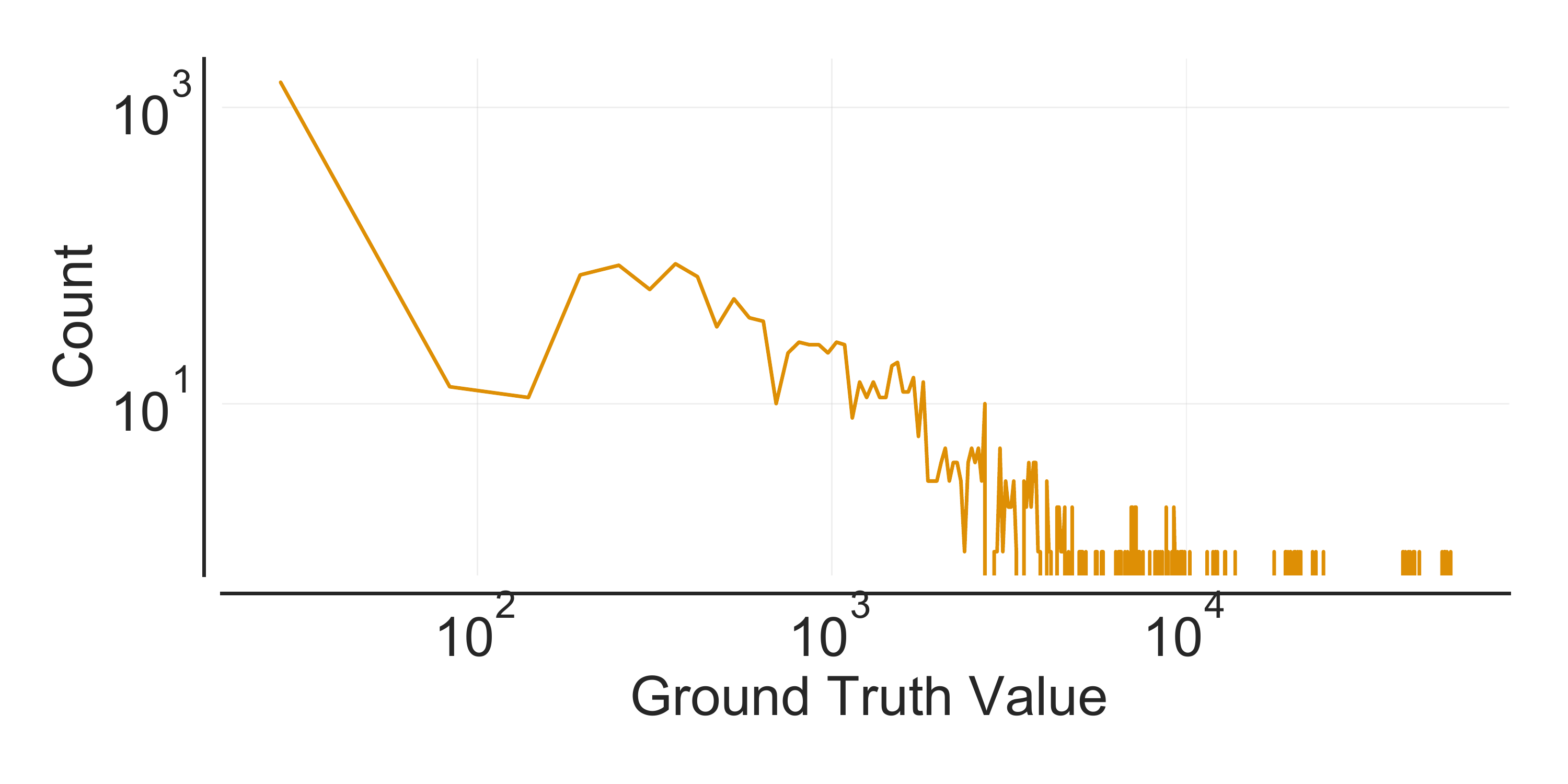}
\includegraphics[width=0.5\columnwidth]{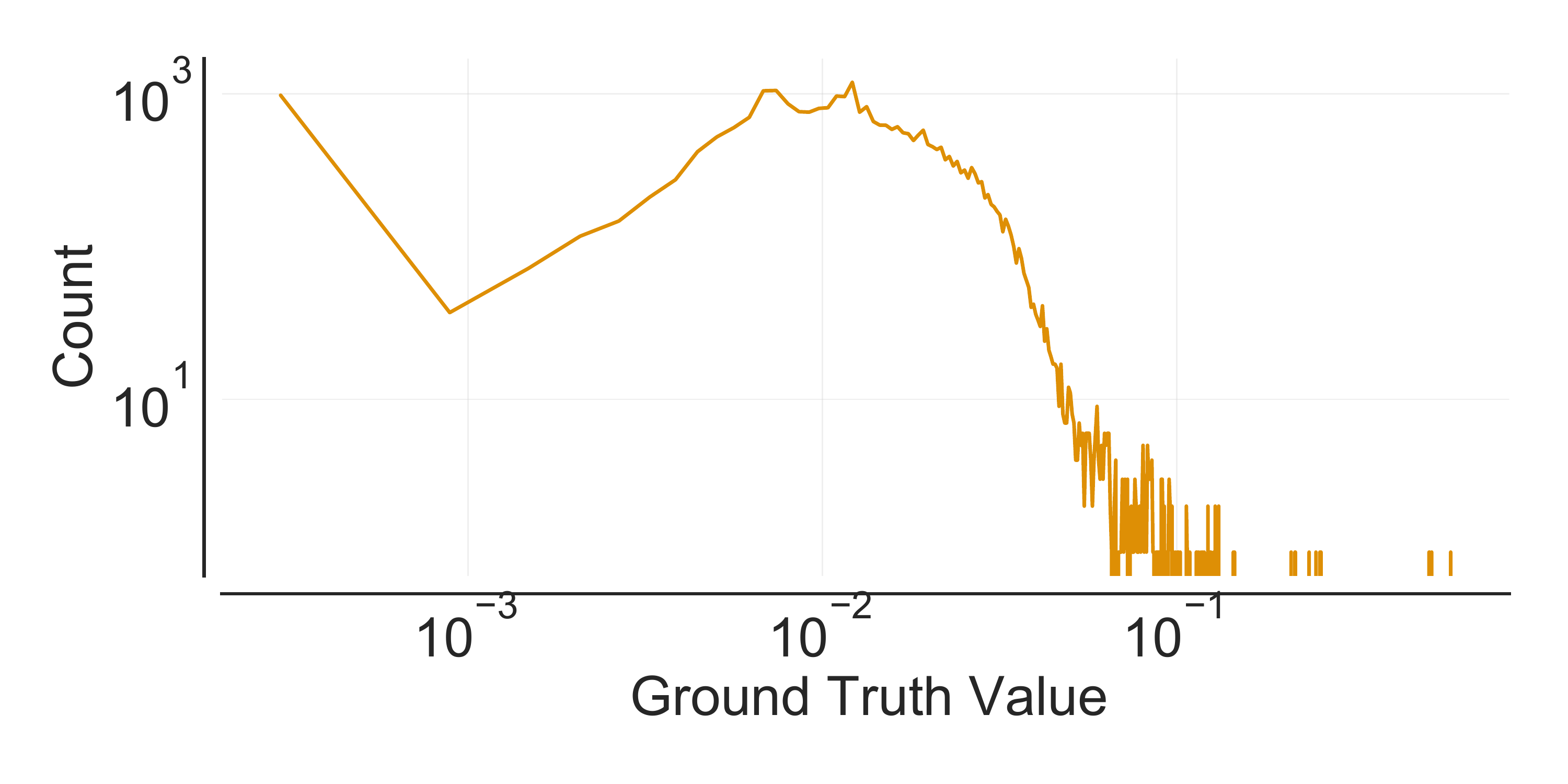}
}
\centerline{
\includegraphics[width=0.5\columnwidth]{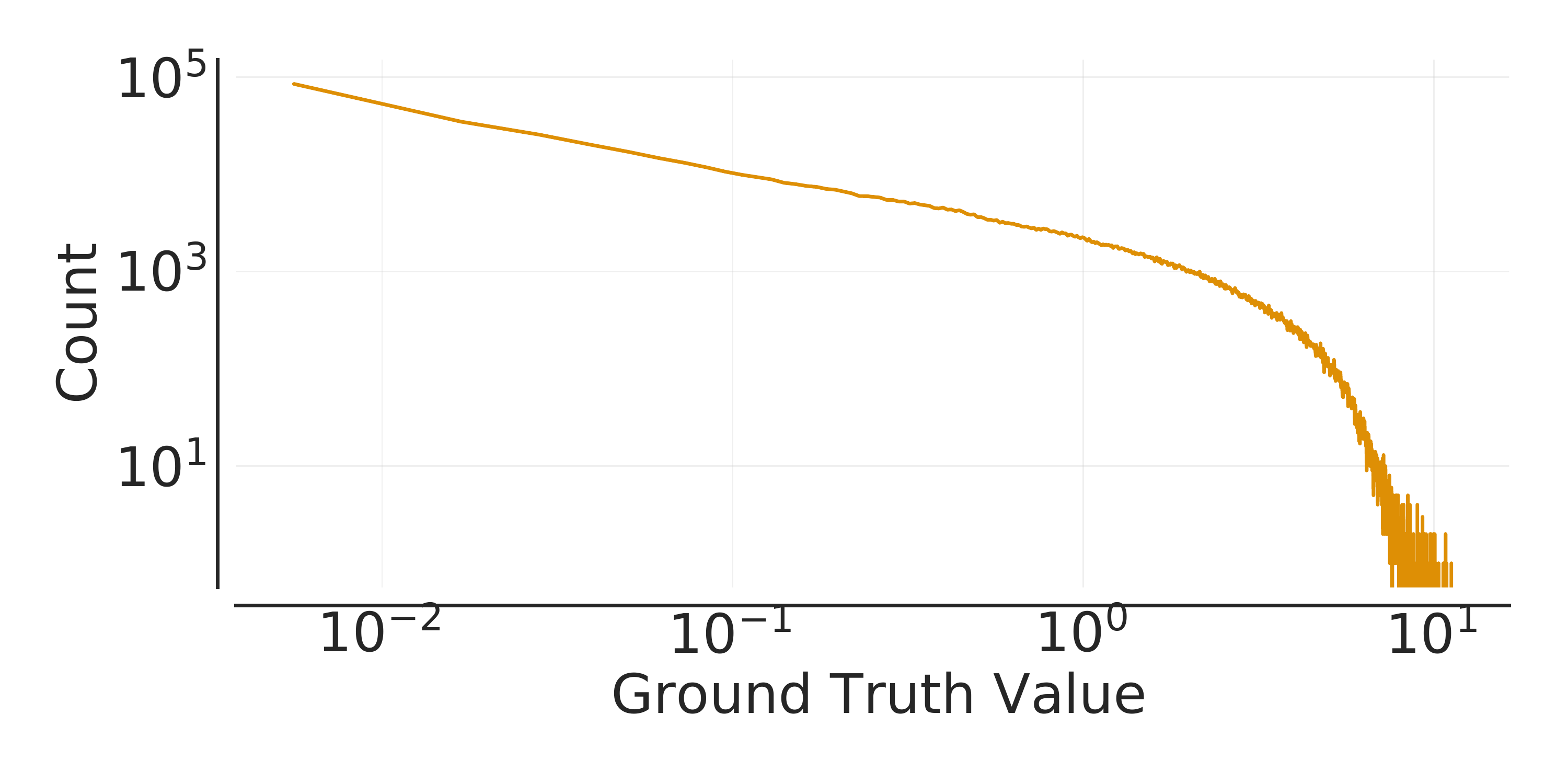} \includegraphics[width=0.5\columnwidth]{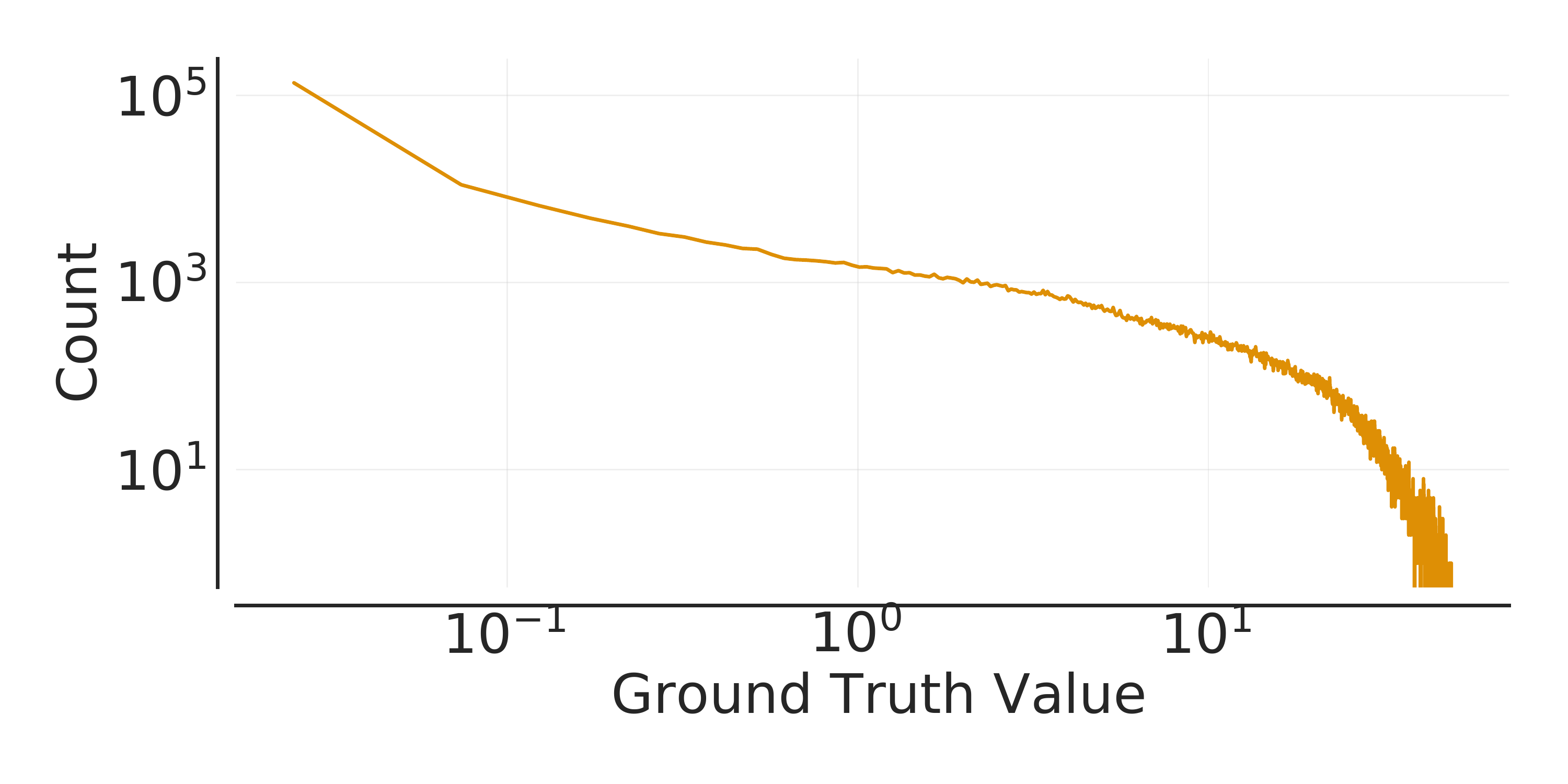}
}
\caption{Log-log plots of distribution of ground truth labels for the time series [Electricity: Top left, Traffic: Top right] and trajectory [ETH-UCY: Bottom left, nuScenes: Bottom right] forecasting datasets. The value for time series datasets represents energy and occupancy and for the trajectory datasets represents normalized 2D coordinates through whole prediction horizon. All datasets exhibit long tail behavior.} 
\label{fig:log_log_data}
\end{center}
\vskip -0.1in
\end{figure}

Long tail distribution of data can be seen in numerous real world datasets. This is evident for the four benchmark forecasting datasets (Electricity~\cite{Dua:2019}, Traffic~\cite{Dua:2019}, ETH-UCY~\cite{pellegrini2009you,lerner2007crowds}, and nuScenes~\cite{caesar2020nuscenes}) studied in this work. We can see the distribution of ground truth values ($\V{y}$) for all of them in Figure \ref{fig:log_log_data}. We use log-log plots  to increase the visibility of the long tail behavior present in the data -- smaller $\V{y}$ values (constituting the minority on a linear scale) occur very frequently, while majority of values are very rare (creating the tail). In addition to the long tail data distribution, we also identify the long tail distribution of forecasting error from deep learning models (such as DeepAR~\cite{salinas2020deepar}, Trajectron++~\cite{salzmann2020trajectron++}, and Trajectron++EWTA~\cite{makansi2019overcoming}) (as seen in Appendix~\ref{section:error_distribution}). 

We hypothesize that long tail behavior in forecasting error distribution originates from the long tail behavior in data distribution, as well as  the nature of gradient based deep learning. Therefore, modifying the loss function to account for the shape of the distribution  would potentially lead to better tail performance. Next, we present two loss functions based on the moment of the error distribution. 

\subsection{Pareto Loss}
\label{section:pareto_loss}

Long tail distributions naturally lend themselves to analysis using Extreme Value Theory (EVT). \cite{mcneil1997estimating} shows that long tail behavior can be modeled as a generalized Pareto distribution (GPD). The probability distribution function (pdf) of the GPD is,
\begin{equation}
f_{(\xi, \eta, \mu)}(a) = \frac{1}{\eta}\left(1 + \xi\left(\frac{a - \mu}{\eta}\right)\right)^{-(\frac{1}{\xi} + 1)}
\label{eq:gpd_pdf}
\end{equation}

where the parameters are location ($\mu$), scale ($\eta$) and shape ($\xi$). The pdf for GPD is defined for $a \geq 0$ when $\xi \geq 0$ and for $0 \leq a \leq - \eta/\xi$ when $\xi < 0$. $\mu$ can be set to 0 without loss of generality as it represents translation along the x axis. We can drop the scaling term $\frac{1}{\eta}$ as the pdf will be scaled using a hyperparameter. The simplified pdf is,
\begin{equation}
f_{(\xi, \eta)}(a) = \left(1 + \frac{\xi a}{\eta}\right)^{-(\frac{1}{\xi} + 1)}
\label{eq:gpd_simple_pdf}
\end{equation}

The high-level idea of Pareto loss is to fit a GPD to the loss distribution to reprioritize the learning of easy and difficult (tail) examples. Let the loss function used by a given machine learning model be denoted as $l$. In probabilistic forecasting,  a commonly used loss is Negative Log Likelihood (NLL) loss:  $l_i = -\log(p(\V{y}^{(i)} | \V{x}^{(i)}))$  where $\langle\V{x}^{(i)}, \V{y}^{(i)}\rangle$ is the $i^{th}$ training example, and $\V{\hat{y}}^{(i)}$ the model prediction. As the pdf in Eq.\eqref{eq:gpd_simple_pdf} only allows non-negative input, the loss $l$ has to be lower-bounded. We propose to use an auxiliary loss $\hat{l}$ to fit the GPD. For NLL which can be unbounded for continuous distributions, the auxiliary loss can simply be Mean Absolute Error (MAE): $\hat{l}_i = \frac{1}{h} \sum_{j=t+1}^{t+h} |\hat{y}_j^{(i)} - y_j^{(i)}|$. 

There are two main classes of methods for modifying loss functions to improve tail performance: regularization \cite{ren2020balanced, makansi2021exposing} and re-weighting \cite{lin2017focal, lu2018deep, yang2021delving}. Both classes are characterized by different behavior on tail data~\cite{ren2020balanced}. Inspired by these, we propose two variations of the Pareto Loss using the distribution fitted on $\hat{l}$: Pareto Loss Margin (PLM) and Pareto Loss Weighted (PLW). 

PLM is based on the principles of margin-based regularization \cite{ren2020balanced, liu2016large} which assigns larger penalties (margins) to harder examples. For a given hyperparameter $\lambda$, PLM is defined as,

\begin{equation}
l_{plm} = l + \lambda * r_{plm}(\hat{l}) 
\label{eq:plm_eqn}
\end{equation}
\noindent where
\begin{equation}
r_{plm}(\hat{l}) = 1 - f_{(\xi, \eta)}(\hat{l})
\label{eq:plm_eqn_r}
\end{equation}
which uses GPD to calculate the additive margin.

An alternative is to reweigh the loss terms using the loss distribution. For a given hyperparameter $\lambda$, PLW is defined as,
\begin{equation}
l_{plw} = w_{plw}(\hat{l}) * l
\label{eq:plw_eqn}
\end{equation}
\noindent where
\begin{equation}
w_{plw}(\hat{l}) = 1 - \lambda * f_{(\xi, \eta)}(\hat{l})
\label{eq:plw_eqn_w}
\end{equation}
which uses GPD to reweigh the loss of each sample.

\subsection{Kurtosis Loss}
Kurtosis measures the tailedness of a distribution as the scaled fourth moment about the mean. To increase the emphasis on tail examples, we use this measure to propose kurtosis loss. For a given hyperparameter $\lambda$ and using the same notation as Sec.\ref{section:pareto_loss} kurtosis loss is defined as,

\vspace{-1mm}
\begin{equation}
l_{kurt} = l + \lambda * r_{kurt}(\hat{l})
\label{eq:kbr}
\end{equation}

\noindent where $r_{kurt}(\hat{l})$ is the contribution of an example to kurtosis for a batch

\begin{equation}
r_{kurt}(\hat{l}) = \left(\frac{\hat{l} - \mu_{\hat{l}}}{\sigma_{\hat{l}}}\right)^4
\label{eq:kbr_e}
\end{equation}

where $\mu_{\hat{l}}$ and $\sigma_{\hat{l}}$ are the mean and standard deviation of the auxiliary loss ($\hat{l}$) values for a batch of examples.

We propose to use the auxiliary loss $\hat{l}$ distribution to compute kurtosis, as performance metrics in forecasting tasks frequently involve versions of L1 or L2 distance such as RMSE, MAE, or ADE. The goal is to decrease the long tail for these metrics, which might not correlate well with the base loss $l$. The example in Sec. \ref{section:pareto_loss} where $l$ is NLL loss and $\hat{l}$ is MAE loss illustrates this requirement well.

Kurtosis loss and pareto loss are related approaches to handling long tail behavior. Pareto Loss is a weighted sum of moments about $0$ while kurtosis loss is the fourth moment about the mean. Let $b = \frac{\xi a}{\eta}$ and $c = -(\frac{1}{\xi} + 1)$, then the Taylor expansion for the GPD pdf from Eq.\eqref{eq:gpd_simple_pdf} is,

\begin{equation}
(1+b)^c = 1 + cb +\frac{c(c-1)}{2!}b^2 + \frac{c(c-1)(c-2)}{3!}b^3 + \cdots
\label{eq:gpd_series}
\end{equation}

For $c < 0$ or equivalently $\xi < -1$ or $\xi > 0$,  the coefficients are positive for even moments and negative for odd moments. Even moments are always symmetric and positive, while odd moments are positive only for right-tailed distributions. Since we use the negative of the pdf, it yields an asymmetric measure of the right tailedness of a value in the distribution.

Kurtosis loss uses the fourth moment about the distribution mean. This is a symmetric and positive measure, but in the context of right tailed distributions, kurtosis serves as a good measure of the long tailedness of the distribution. GPD and kurtosis are visualised in Appendix \ref{section:pareto_kurtosis}


\section{Experiments}
We evaluate our methods on two probabilistic forecasting tasks: time series forecasting and trajectory prediction.

\subsection{Setup}

\paragraph{Datasets.}
For time series forecasting, we use electricity and traffic datasets from the UCI ML repository \cite{Dua:2019} used in \cite{salinas2020deepar} as benchmarks. We also generate three synthetic 1D time series datasets, Sine, Gaussian and Pareto, to further our understanding of long tail behavior. 

For trajectory prediction, we use two benchmark datasets: a pedestrian trajectory dataset ETH-UCY (which is a combination of ETH~\cite{pellegrini2009you} and UCY~\cite{lerner2007crowds} datasets) and a vehicle trajectory dataset nuScenes ~\cite{caesar2020nuscenes}. Details regarding the datasets are available in Appendix \ref{section:dataset_description}.

\paragraph{Baselines.}
We compare with the following baselines representing SoTA in long tail mitigation for different tasks:

\begin{itemize}[itemsep=1mm, topsep=0mm]
    \item Contrastive Loss:~\cite{makansi2021exposing} uses contrastive loss as a regularizer to group examples together based on Kalman filter prediction errors.

    \item Focal Loss:~\cite{lin2017focal} uses L1 loss to reweigh loss terms.

    \item Shrinkage Loss:~\cite{lu2018deep} uses a sigmoid-based function to reweigh loss terms.

    \item Label Distribution Smoothing (LDS):~\cite{yang2021delving} uses symmetric kernel to smooth the label distribution and use its inverse to reweigh loss terms.
\end{itemize}

Focal Loss, Shrinkage Loss, and LDS were originally proposed for classification and/or regression and required adaptation in order to be applicable to the forecasting task. For details on baseline adaptation, please see Appendix~\ref{section:baseline_adapt}.

\paragraph{Evaluation Metrics.}

We use two common metrics for the evaluation of trajectory prediction models: Average Displacement Error (ADE), which is the average L2 distance between total predicted trajectory and ground truth, and Final Displacement Error (FDE) which is the L2 distance for the final timestep. For time series forecasting, we use Normalized Deviation (ND) and Normalized Root Mean Squared Error (NRMSE).

Apart from the above-mentioned average performance metrics, we introduce metrics to capture performance on the tail. To measure the performance at tail of the distribution, we propose to adapt the Value-at-Risk (VaR Eq.~\eqref{eq:var}) metric:
\begin{equation}
\mathrm{VaR}_{\alpha}(E) = \inf \{e\in E: P(E\geq e)\leq 1-\alpha\} 
\label{eq:var}
\end{equation}

VaR at level $\alpha \in (0,1)$ is the smallest error $e$ such that the probability of observing error larger than $e$ is smaller than $1-\alpha$, where $E$ is the error distribution. This evaluates to the $\alpha^{th}$ quantile of the error distribution. We propose to measure VaR at three different levels: $0.95$, $0.98$, and $0.99$. 

In addition, we use skew, kurtosis, and max error to further assess the tail performance. Skew and Kurtosis as metrics are meaningful only when looked at in conjunction with the mean. A distribution with a higher mean and lower skew and kurtosis does not imply a less severe tail.
\begin{figure}[t!]
\vskip 0.1in
\centerline{
\includegraphics[width=0.6\columnwidth]{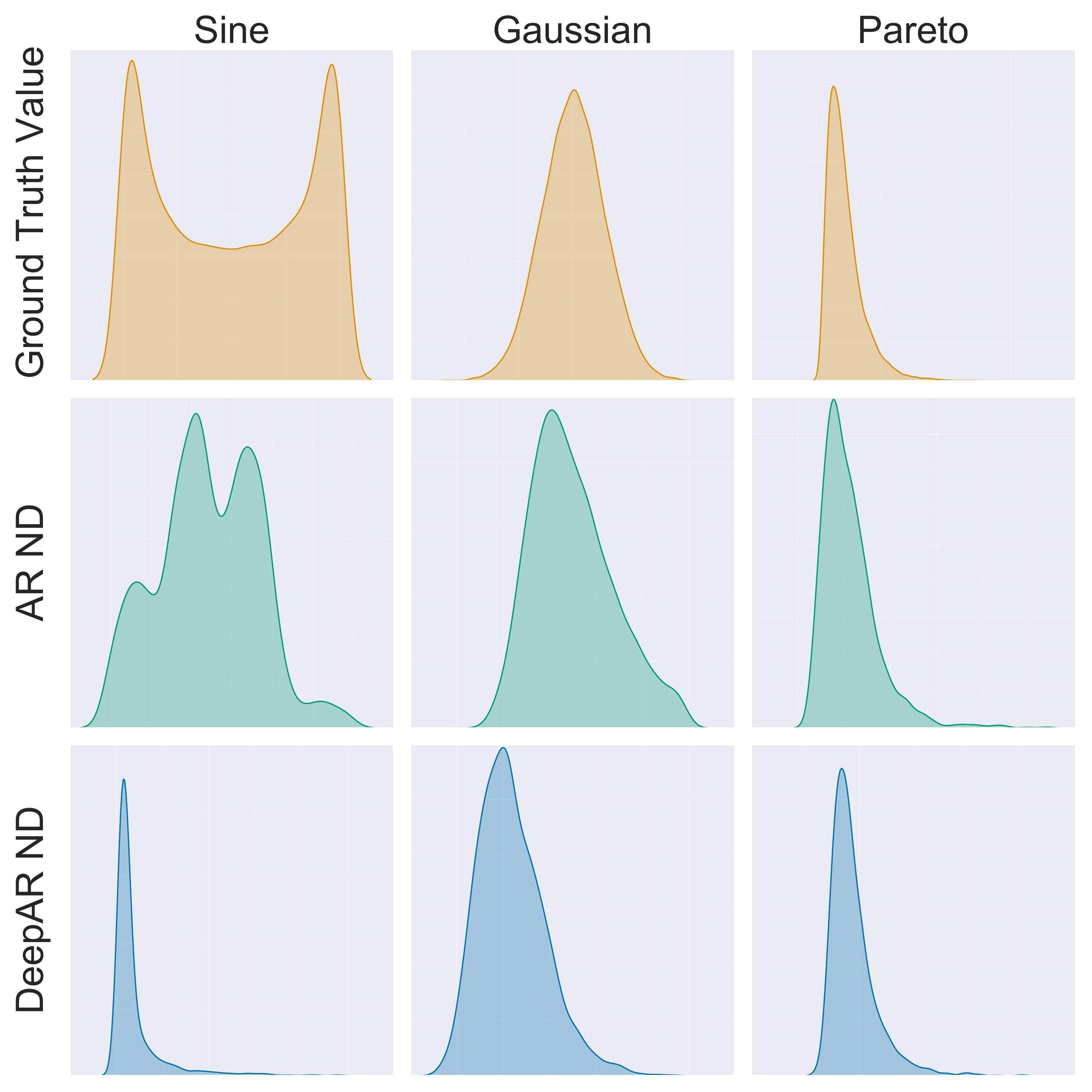}
}
\caption{Top Row: Ground truth distribution for synthetic datasets. Middle Row: Normalized Deviation (ND) error distribution for prediction using AutoRegression. Bottom Row : ND error distribution for prediction using DeepAR. Datasets (L to R): Sine, Gaussian, Pareto. The error distribution for AR and DeepAR on these datasets indicate that long tail behavior in error originates due to both long tail data distribution and gradient learning. \textbf{Note}: the x-axes for plots in the same column or y-axes for plots in the same row are not on the same range of values}
\label{fig:synthetic_datasets}
\end{figure}

\subsection{Synthetic Dataset Experiments}
\label{section:syn_dataset_exp}

In order to better understand the long tail error distribution, we perform  experiments on three synthetic datasets. The task is to forecast 8 steps ahead given a history of 8 time steps. We use AutoRegression (AR) and DeepAR \cite{salinas2020deepar} as forecasting models to perform this task. The top row in Figure \ref{fig:synthetic_datasets} shows that among the datasets, only Gaussian and Pareto show tail behavior in the data distribution. Pareto dataset in particular is the only one to display long tail behavior. AR and DeepAR have different error distribution across the datasets. Based on these results, we make the following hypotheses for the sources of long-tailedness.

\textit{Source 1: Long Tail in Data.} The data distributions for Gaussian and Pareto datasets have similar tail behavior to the error distribution for both models, AR and DeepAR. This indicates that the long tail in data is a likely cause of long tail behavior in error. This connection is also well established as class imbalance for classification tasks~\cite{van2018inaturalist, liu2019large}.

\textit{Source 2: Deep Learning Model.} The results on the Sine dataset illustrate that even in the absence of long tail in the data, we can have long tail in the error distribution. The AR model, however, does not show long tail behavior for error. This indicates that the observed long tail behavior in error for DeepAR is model induced. We hypothesize that this is caused by DeepAR overfitting to simpler examples due to the nature of gradient based learning. Further results and analysis on these datasets can be found in Appendix \ref{section:syn_dataset_analysis}.

The difference between AR and DeepAR error distributions also suggests that assuming tail overlap between deep learning and non-deep learning methods (such as Kalman filter used by~\cite{makansi2021exposing}) might not generalize well.


\begin{table*}[t!]
\vspace{-0.02in}
\caption{Performance on the \textbf{Electricity Dataset} (ND/NRMSE). PLW, PLM and Kurtosis (Ours) all improve on the average as well as tail metrics. Baseline methods perform slightly worse on average as compared to DeepAR.  Results indicated as \better{Better} and \best{Best}}
\label{table:deepar_elect}
\begin{center}
\begin{small}
\begin{sc}
\begin{tabular}{lcccccccc}
\toprule
Method & Metric & Mean$\downarrow$ & $\mathrm{VaR}_{95}\downarrow$ & $\mathrm{VaR}_{98}\downarrow$ & $\mathrm{VaR}_{99}\downarrow$ & Max$\downarrow$ & Kurtosis$\downarrow$ & Skew$\downarrow$ \\
\midrule
DeepAR & ND & 0.0584 & \best{0.0796} & 0.2312 & 0.4429 & 4.1520 & 426.5906 & 18.4057\\
& NRMSE & 0.2953 & \best{0.0972} & 0.2595 & 0.5263 & 5.4950 & 470.8968 & 19.4827\\
\midrule
+ Contrastive Loss & ND & 0.0618 & 0.0872 & \better{0.2102} & \better{0.4274} & \better{4.0004} & \better{384.568} & \better{17.5604}\\
& NRMSE & 0.3062 & 0.1069 & \better{0.2481} & 0.5392 & \better{5.1606} & \better{415.3592} & \better{18.3051}\\
\midrule
+ Focal Loss & ND & 0.0628 & 0.0853 & 0.2694 & \better{0.4398} & 4.3263 & \better{412.5172} & \better{18.0739}\\
& NRMSE & 0.3139 & 0.1052 & 0.3137 & 0.5297 & 5.7797 & \better{469.7605} & \better{19.3916}\\
\midrule
+ Shrinkage Loss & ND & 0.0694 & 0.0956 & 0.2334 & 0.4446 & 4.4714 & \better{325.7401} & \better{16.3852}\\
& NRMSE & 0.3244 & 0.1156 & 0.2828 & \better{0.5177} & \better{5.4245} & \best{336.7777} & \better{16.5656}\\
\midrule
+ LDS & ND & 0.0634 & 0.0890 & \better{0.2238} & 0.4925 & \better{3.8625} & \better{335.2523} & \better{16.2944}\\
& NRMSE & \better{0.2923} & 0.1149 & 0.2787 & 0.5458 & \better{4.9234} & \better{373.4702} & \better{17.1249}\\
\midrule
+ Kurtosis Loss (Ours) & ND & \better{0.0567} & 0.0842 & \better{0.2151} & \better{0.4120} & \best{3.2738} & \best{300.3517} & \best{15.4597}\\
& NRMSE & \best{0.2631} & 0.1046 & \better{0.2732} & \best{0.4779} & \best{4.2613} & \better{339.3773} & \best{16.4892}\\
\midrule
+ PLM (Ours) & ND & \best{0.0564} & 0.0799 & \best{0.1900} & \better{0.4164} & \better{3.4576} & \better{359.6645} & \better{16.9243}\\
& NRMSE & \better{0.2783} & 0.1000 & \best{0.2343} & \better{0.5102} & \better{4.7494} & \better{423.2319} & \better{18.3994}\\
\midrule
+ PLW (Ours) & ND & \better{0.0578} & \best{0.0796} & \better{0.2121} & \best{0.3558} & \better{3.4647} & \better{329.0847} & \better{16.393}\\
& NRMSE & \better{0.2807} & 0.0984 & \better{0.2555} & \better{0.4809} & \better{4.6040} & \better{366.6818} & \better{17.3120}\\
\bottomrule
\end{tabular}
\end{sc}
\end{small}
\end{center}
\vskip -0.1in
\end{table*}
\begin{table*}[t!]
\vspace{-0.02in}
\caption{Performance on the \textbf{Traffic Dataset} (ND/NRMSE). Pareto Loss Margin (Ours) delivers best overall results improving on both average and the tail. Regularization methods in general fare better than weighting methods due to a very long tail. Among the baseline methods contrastive loss exhibits most consistent improvements. Results indicated as \better{Better} and \best{Best}}
\label{table:deepar_traffic}
\begin{center}
\begin{small}
\begin{sc}
\begin{tabular}{lcccccccc}
\toprule
Method & Metric & Mean$\downarrow$ & $\mathrm{VaR}_{95}\downarrow$ & $\mathrm{VaR}_{98}\downarrow$ & $\mathrm{VaR}_{99}\downarrow$ & Max$\downarrow$ & Kurtosis$\downarrow$ & Skew$\downarrow$\\
\midrule
DeepAR & ND & 0.1741 & \best{0.6866} & 25.5840 & 32.1330 & 84.1582 & 41.2804 & 6.1700\\
& NRMSE & \best{0.4465} & \best{1.2283} & 6.0283 & 7.5988 & 18.8103 & 37.0089 & 5.7343\\
\midrule
+ Contrastive Loss & ND & 0.2052 & 0.7463 & \best{24.3737} & \better{30.5117} & \better{81.1716} & 42.1391 & 6.2282\\
& NRMSE & 0.4667 & 1.2956 & \better{5.7747} & \better{7.2342} & \better{18.3360} & \better{36.4420} & \better{5.6834}\\
\midrule
+ Focal Loss & ND & 0.4903 & 1.1553 & 26.7537 & \best{30.1506} & \best{52.8272} & \best{28.5912} & \best{5.4325}\\
& NRMSE & 0.7302 & 1.6485 & 6.5880 & \better{7.3660} & \better{13.7985} & \best{24.6181} & \best{4.9104}\\
\midrule
+ Shrinkage Loss & ND & 0.2431 & 0.8380 & \better{25.3381} & 32.9147 & 85.2713 & 45.0172 & 6.3935\\
& NRMSE & 0.5114 & 1.3099 & 6.0418 & 7.8882 & 19.0771 & 39.5592 & 5.8742\\
\midrule
+ LDS & ND & 0.4763 & 1.4781 & 28.9162 & 38.4263 & 126.5733 & 49.2714 & 6.5445\\
& NRMSE & 0.7829 & 1.8702 & 6.8826 & 9.2061 & 27.3684 & 39.8322 & \better{5.7109}\\
\midrule
+ Kurtosis Loss (Ours) & ND & 0.2022 & 0.7653 & \better{25.3752} & \better{31.4677} & \better{62.9173} & \better{35.298} & 5.8785\\
& NRMSE & 0.4892 & 1.4072 & \better{6.0263} & \better{7.3369} & \best{13.7783} & \better{29.6338} & 5.2683\\
\midrule
+ PLM (Ours) & ND & \best{0.1594} & 0.7115 & \better{24.5911} & \better{30.331} & 90.3169 & 42.5373 & 6.1829\\
& NRMSE & 0.4600 & 1.3881 & \best{5.6779} & \best{7.0033} & 20.5736 & \better{36.7518} & \better{5.6005}\\
\midrule
+ PLW (Ours) & ND & 0.3751 & 1.0495 & \better{25.4471} & \better{31.6621} & \better{65.759} & \better{35.4836} & \better{5.8813}\\
& NRMSE & 0.6238 & 1.4914 & 6.0552 & \better{7.3491} & \better{13.8938} & \better{28.9214} & \better{5.1844}\\
\bottomrule
\end{tabular}
\end{sc}
\end{small}
\end{center}
\vskip -0.1in
\end{table*}


\begin{table*}[t!]
\caption{Macro-averaged performance on the \textbf{ETH-UCY Dataset} (ADE/FDE). Our approaches improve tail performance better than existing baselines. The improvements are most significant for far-future prediction (FDE). PLM improves well across prediction horizon (ADE). Results indicated as \better{Better} and \best{Best}.}
\label{table:eth_ucy_20guess}
\vspace{-0.01in}
\begin{center}
\begin{small}
\begin{sc}
\begin{tabular}{lccccccccc}
\toprule
Method & Mean$\downarrow$ & $\mathrm{VaR}_{95}\downarrow$ & $\mathrm{VaR}_{98}\downarrow$ & $\mathrm{VaR}_{99}\downarrow$ & Max$\downarrow$ & Kurtosis$\downarrow$ & Skew$\downarrow$\\
\midrule
Traj++ & 0.21/0.41 & 0.56/1.33 & 0.78/1.97 & 0.98/2.47 & 2.33/5.04 & 16.02/16.09 & 3.02/3.26\\
\midrule
Traj++EWTA & \best{0.16}/0.33 & 0.43/1.05 & 0.60/1.53 & 0.76/1.89 & 1.63/3.95 & 16.40/19.21 & 2.97/3.34\\
 + Contrastive & 0.17/0.34 & 0.43/\better{1.03} & 0.62/1.56 & 0.79/1.89 & 1.67/4.02 & \better{16.37}/\better{18.51} & \better{2.96}/3.35\\
 + Focal Loss & \best{0.16}/\better{0.32} & \better{0.40}/\better{0.89} & \better{0.54}/\better{1.28} & \better{0.66}/\better{1.57} & \better{1.50}/\better{3.50} & \better{14.95}/\better{17.80} & \better{2.74}/\better{3.18}\\
 + Shrinkage Loss & \best{0.16}/0.33 & 0.43/1.05 & \better{0.58}/\better{1.50} & \better{0.74}/\better{1.84} & 1.66/3.95 & 16.67/19.54 & 3.00/3.41\\
 + LDS & 0.17/0.35 & 0.44/\better{1.04} & \better{0.57}/\better{1.45} & 0.78/\better{1.86} & 1.69/\better{3.85} & 19.80/\better{19.12} & 3.18/3.39\\
 + Kurtosis Loss (ours) & 0.17/0.34 & 0.46/\better{0.98} & \better{0.59}/\better{1.25} & \better{0.67}/\better{1.47} & \best{1.22}/\best{2.77} & \best{5.28}/\best{7.25} & \better{1.77}/\better{2.11} \\
 + PLM (ours) & \best{0.16}/\best{0.30} & \best{0.38}/\best{0.81} & \best{0.52}/\better{1.20} & \best{0.63}/\better{1.49} & \better{1.30}/\better{3.20} & \better{12.01}/\better{16.90} & \better{2.41}/\better{3.04}\\
 + PLW (ours) & 0.21/0.36 & 0.46/\better{0.84} & \better{0.55}/\best{1.08} & \best{0.63}/\best{1.32} & \better{1.25}/\better{2.93} & \better{6.62}/\better{10.52} & \best{1.69}/\best{2.08}\\
\bottomrule
\end{tabular}
\end{sc}
\end{small}
\end{center}
\vskip -0.1in
\end{table*}

\begin{table*}[h!]
\caption{Average performance on the \textbf{nuScenes Dataset} (ADE/FDE). Our approaches improve tail performance for far-future prediction (FDE) better than existing baselines. Results indicated as \better{Better} and \best{Best}.}
\label{table:nuscenes_20guess}
\vspace{-0.01in}
\begin{center}
\begin{small}
\begin{sc}
\begin{tabular}{lccccccccc}
\toprule
Method & Mean$\downarrow$ & $\mathrm{VaR}_{95}\downarrow$ & $\mathrm{VaR}_{98}\downarrow$ & $\mathrm{VaR}_{99}\downarrow$ & Max$\downarrow$ & Kurtosis$\downarrow$ & Skew$\downarrow$\\
\midrule
Traj++ & 0.23/0.42 & 0.73/1.62 & 1.11/2.73 & 1.46/3.61 & 7.87/10.98 & 37.74/26.96 & 4.23/4.18\\
\midrule
Traj++EWTA & \best{0.19}/0.34 & 0.65/1.49 & 1.00/2.49 & 1.32/3.34 & 7.07/11.42 & 55.26/36.33 & 5.12/4.88\\
 + Contrastive & \best{0.19}/0.35 & 0.65/1.51 & 1.01/2.58 & 1.36/3.46 & \better{6.82}/\better{10.48} & \better{52.62}/\better{32.32} & \better{5.07}/\better{4.71}\\
 + Focal Loss & \best{0.19}/\better{0.33} & \best{0.56}/\better{1.09} & \better{0.85}/\better{1.95} & \better{1.11}/\better{2.65} & \better{6.55}/11.71 & 60.48/53.60 & 5.14/5.55\\
 + Shrinkage Loss & \best{0.19}/\best{0.32} & \better{0.62}/\better{1.32} & \better{0.96}/\better{2.31} & \better{1.25}/\better{3.17} & \better{6.39}/\better{10.26} & \better{53.5}/36.91 & \better{5.00}/4.95\\
 + LDS & \best{0.19}/\best{0.32} & \better{0.62}/\better{1.26} & \better{0.94}/\better{2.23} & \better{1.20}/\better{2.99} & \best{5.20}/\better{10.53} & \better{46.71}/40.00 & \better{4.75}/5.08\\
 + Kurtosis Loss (ours) & 0.20/0.38 & 0.65/\better{1.35} & \better{0.85}/\better{1.82} & \better{1.03}/\better{2.27} & \better{5.39}/\best{7.52} & \best{28.32}/\best{17.88} & \best{3.24}/\best{3.00}\\
 + PLM (ours) & \best{0.19}/\better{0.33} & \better{0.62}/\better{1.32} & \better{0.95}/\better{2.31} & \better{1.25}/\better{3.18} & \better{6.10}/\better{10.96} & \better{46.43}/37.63 & \better{4.71}/4.96\\
 + PLW (ours) & 0.24/0.37 & \better{0.60}/\best{1.00} & \best{0.82}/\best{1.49} & \best{1.01}/\best{2.01} & 7.51/\better{9.91} & 62.85/42.87 & \better{4.46}/\better{4.57}\\

\bottomrule
\end{tabular}
\end{sc}
\end{small}
\end{center}
\vskip -0.1in
\end{table*}

\subsection{Real-World Experiments}
\label{sec:task_setup}

\paragraph{Time Series Forecasting}
We present average and tail metrics on ND and NRMSE for the time series forecasting task on electricity and traffic datasets in Tables ~\ref{table:deepar_elect} and~\ref{table:deepar_traffic} respectively. We use DeepAR \cite{salinas2020deepar}, one of the SoTA in probabilistic time series forecasting, as the base model. The task for both datasets is to use a  1-week history (168 hours) to forecast for 1 day (24 hours) at an hourly frequency. DeepAR exhibits long tail behavior in error on both datasets (refer Appendix~\ref{section:error_distribution}). The tail of the error distribution is significantly longer for the electricity dataset as compared to the traffic dataset.

\paragraph{Trajectory Forecasting}

We present experimental results on ETH-UCY and nuScenes datasets in Tables~\ref{table:eth_ucy_20guess} and~\ref{table:nuscenes_20guess} respectively. Following~\cite{salzmann2020trajectron++} and~\cite{makansi2021exposing} we calculate model performance based on the best out of 20 guesses. On both datasets we compare our approaches with current SoTA long-tail baseline methods using Trajectron++EWTA~\cite{makansi2021exposing} as a base model due to its SoTA average performance on these datasets. We include the Trajectron++~\cite{salzmann2020trajectron++} results for reference as the previous state-of-the-art method to add a meaningful comparison to the magnitude of performance change obtained by each long tail method. 

On performing a comparative analysis of tail lengths between datasets, we notice that trajectory datasets manifest shorter tails compared to 1D time series datasets. Our Pareto approaches work better on longer tails and for this reason we augment weight and margin for PLM and PLW with an additional Mean Squared Error weight term to internally elongate the tail during the training process.

\subsection{Results Analysis}

\paragraph{Cross-task consistency} As shown in Tables~\ref{table:eth_ucy_20guess} and~\ref{table:nuscenes_20guess}, our proposed approaches, kurtosis loss and PLM, are the only methods improving on tail metrics across all tasks while maintaining the average performance of the base model. Our tasks differ in representation (1D, 2D), severity of long-tail, base model loss function (GaussNLL, EWTA) and prediction horizon. This indicates that our methods generalize to diverse situations better than existing long-tail methods.

\paragraph{Long-tailedness across datasets}

Using Eq.~\eqref{eq:tail_severity} as an indicative measure of the long-tailedness in error distribution, we establish the datasets as ETH-UCY, nuScenes, electricity, and traffic in long-tailedness for the base model (Details in Appendix~\ref{section:tail_severity}). We notice the connections between long-tailedness of the dataset and the performance of different methods.
\vspace{-0.5mm}
\begin{equation}
\mathrm{TailLength}=\frac{\mathrm{VaR}_{95}}{\mathrm{Mean}} + \frac{\mathrm{VaR}_{98}}{\mathrm{VaR}_{95}} + \frac{\mathrm{VaR}_{99}}{\mathrm{VaR}_{98}} + \frac{\mathrm{Max}}{\mathrm{VaR}_{99}}
\label{eq:tail_severity}
\end{equation}

\paragraph{Re-weighting vs Regularization.}

As  mentioned in Section~\ref{section:pareto_loss}, we can categorize loss modifying methods into two classes: re-weighting  (focal loss, shrinkage loss, LDS and PLW)  and regularization (contrastive loss, PLM and kurtosis loss). Re-weighting  multiplies the loss for more difficult examples with higher weights. Regularization  adds higher regularization values for examples with higher loss.

We notice that re-weighting methods perform worse as the long-tailedness increases. In scenarios with longer tails, the weights of tail samples can be very high. Over-emphasizing tail examples hampers the learning for other samples. Shrinkage loss with a bounded weight limits this issue but fails to show tail improvements in longer tail scenarios. PLW is the best re-weighting method on most datasets, likely due to its bounded weights. Inconsistency in average performance is likely due to re-weighting nature of the loss which limits its applicability.

In contrast, regularization methods perform consistently across all tasks both on the tail and average metrics. The additive nature of regularization limits the adverse impact tail samples can have on the learning. This enables these methods to handle different long-tailednesses without degrading the average performance.

\begin{figure*}[t!]
\vskip 0.05in
\begin{center}
\centerline{
\includegraphics[width=0.25\textwidth]{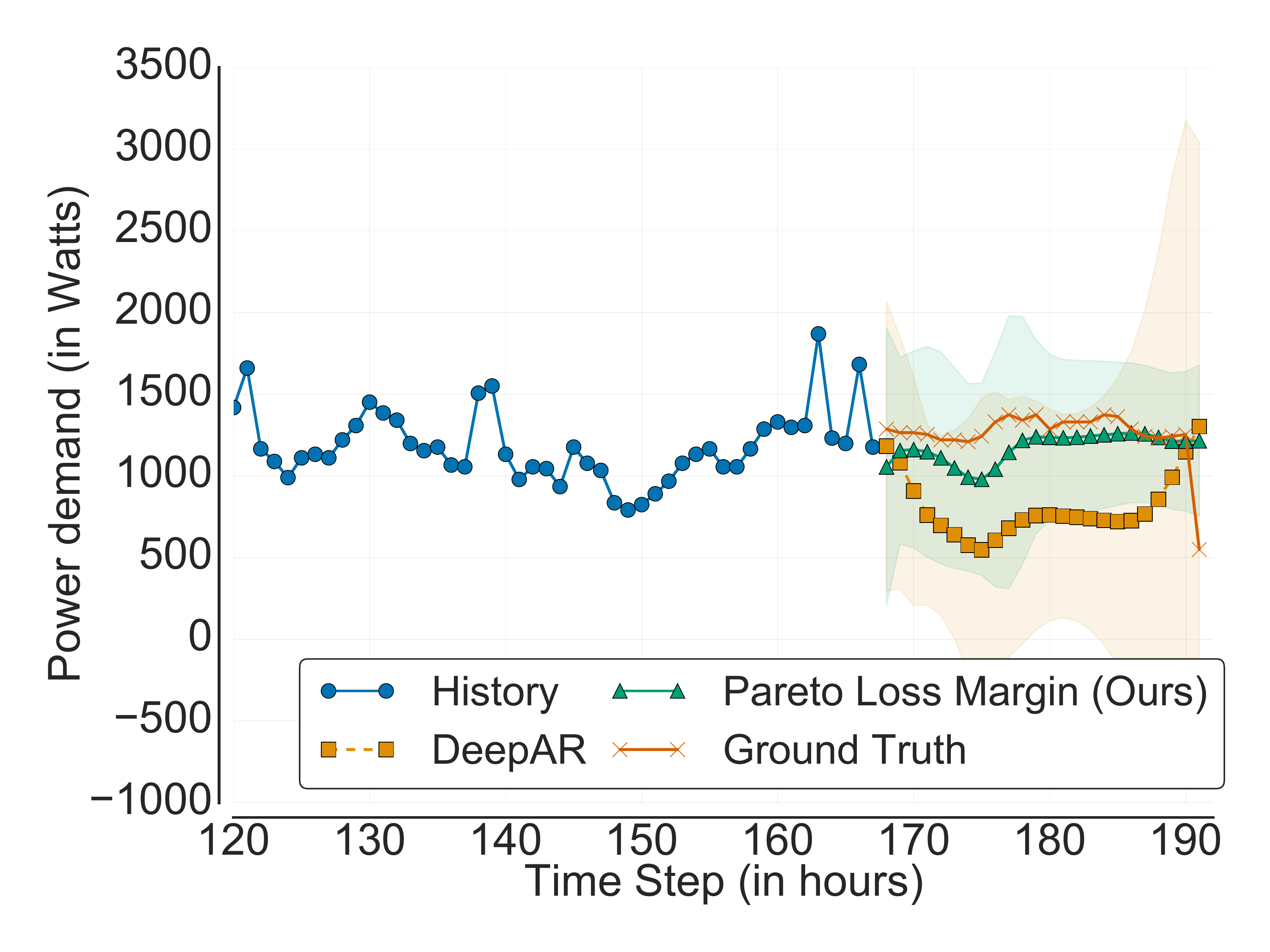}
\includegraphics[width=0.25\textwidth]{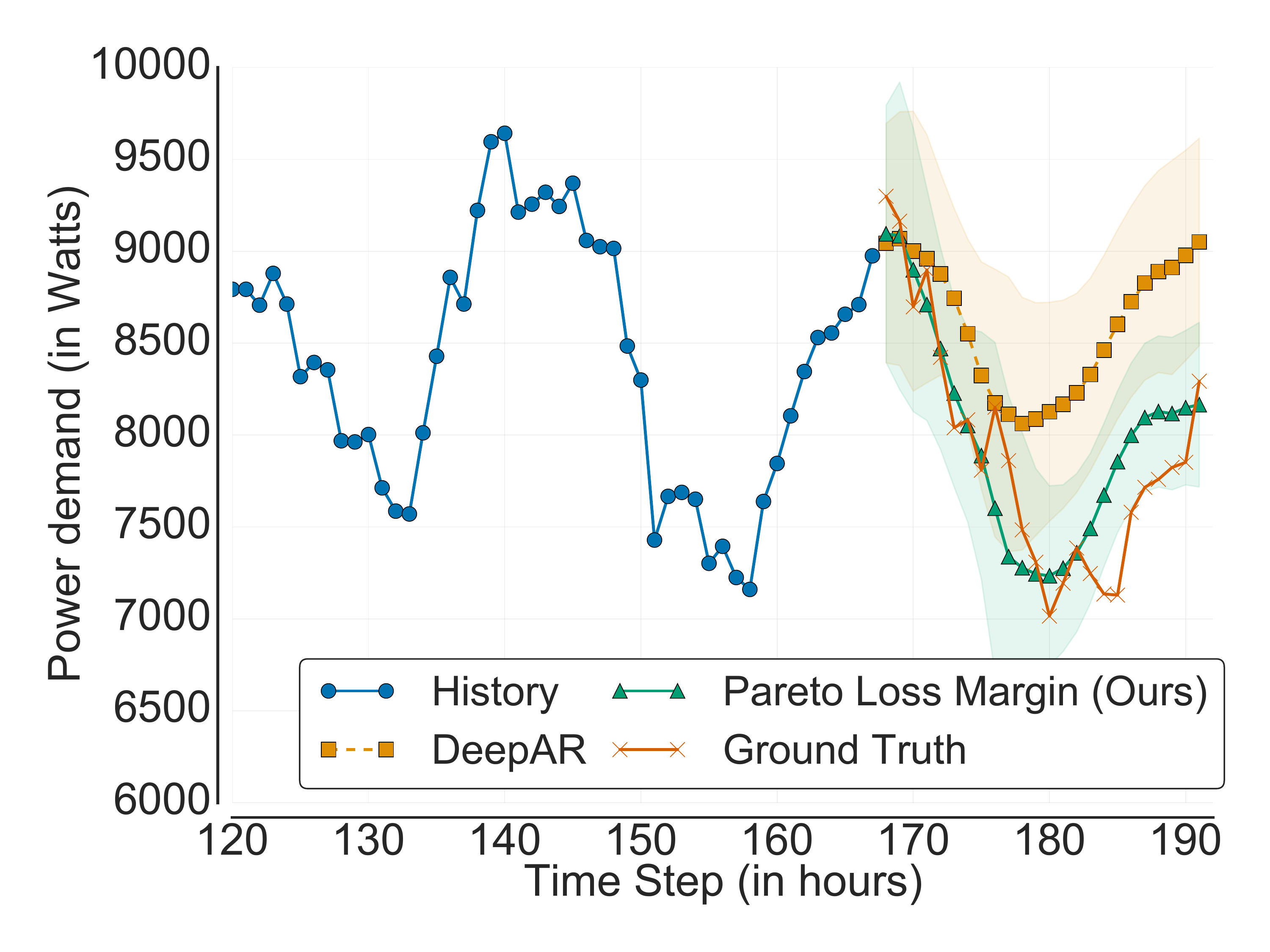}
\includegraphics[width=0.25\textwidth]{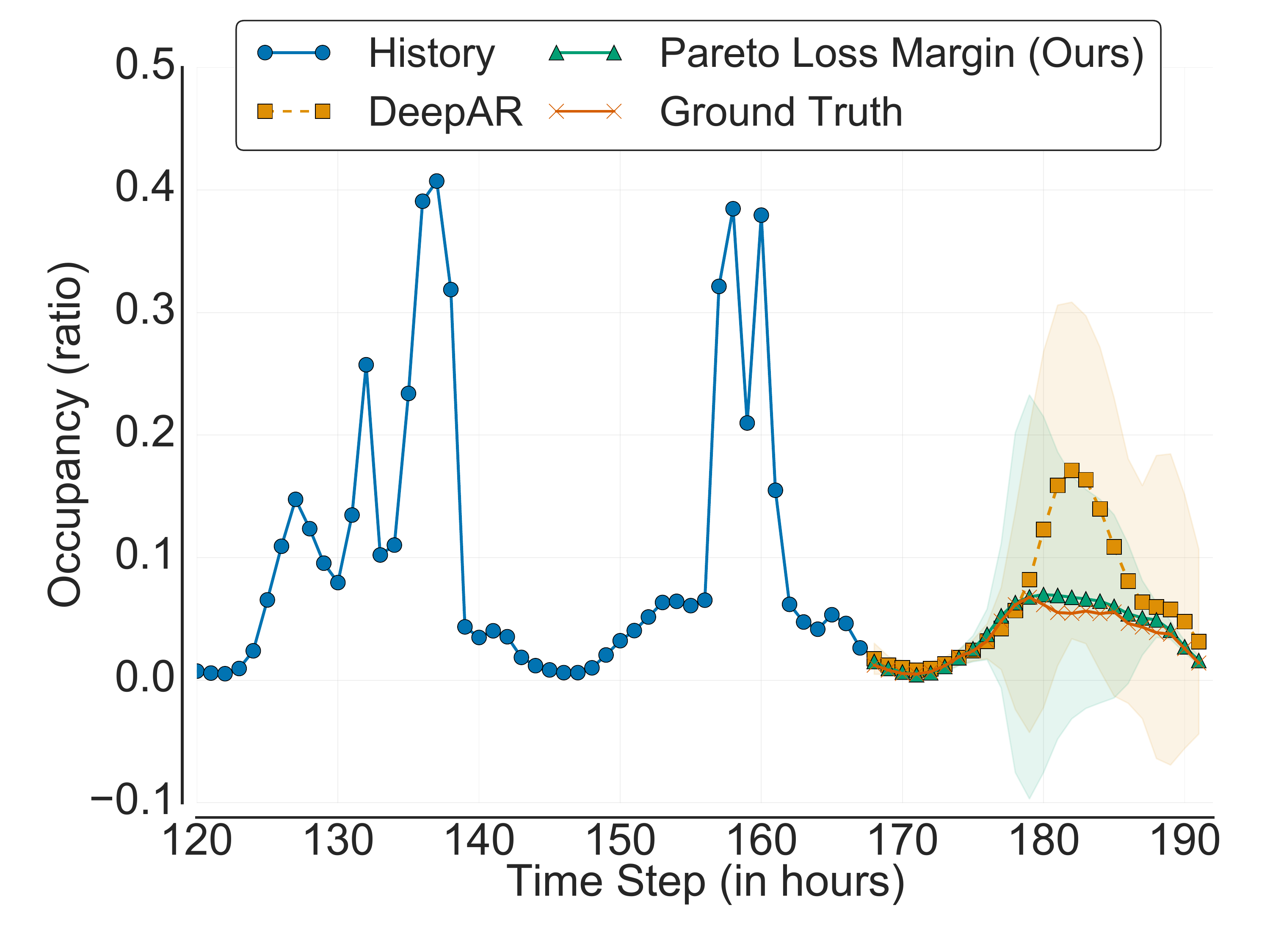}
\includegraphics[width=0.25\textwidth]{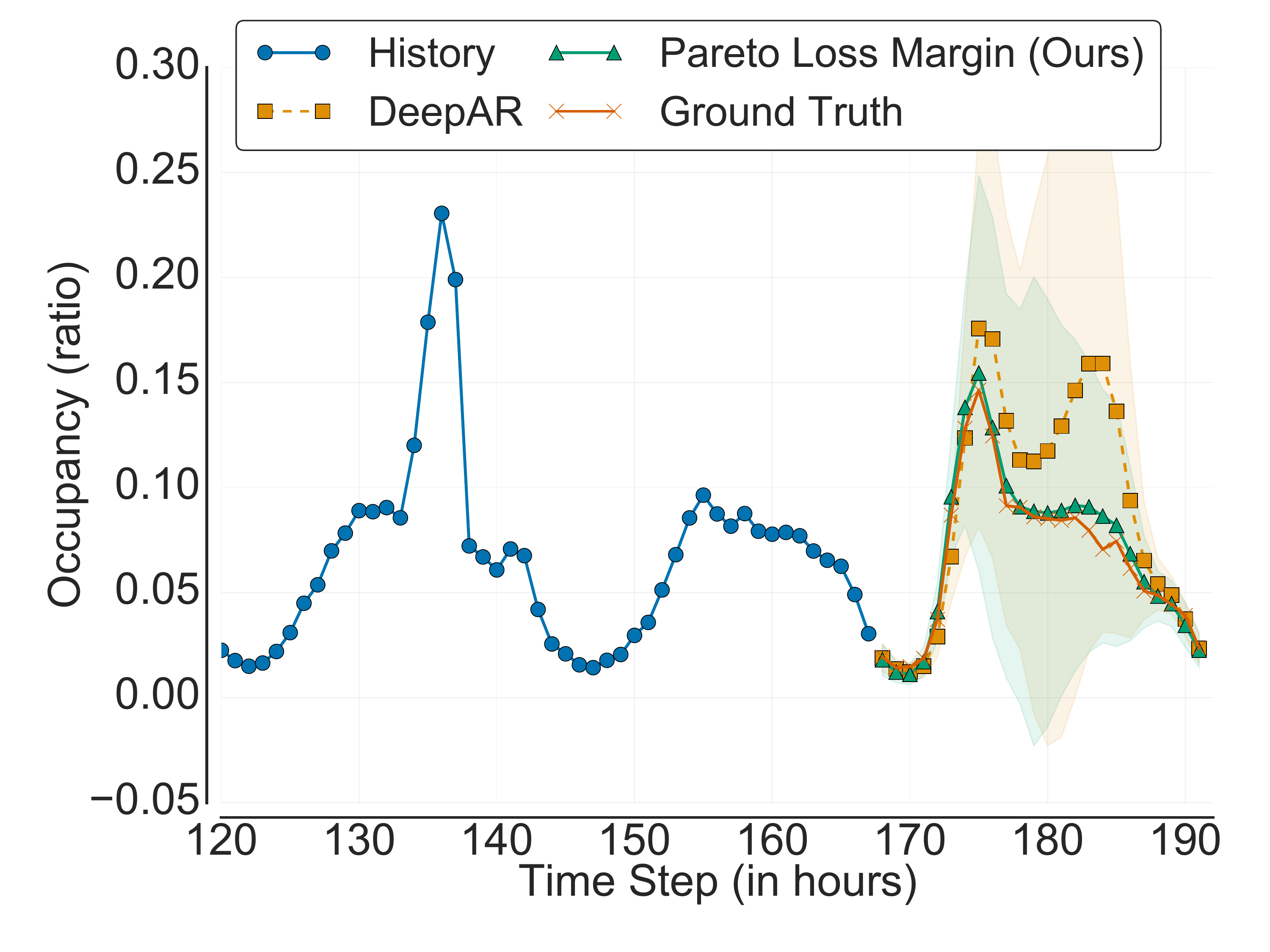}}

\centerline{
\includegraphics[width=0.25\textwidth]{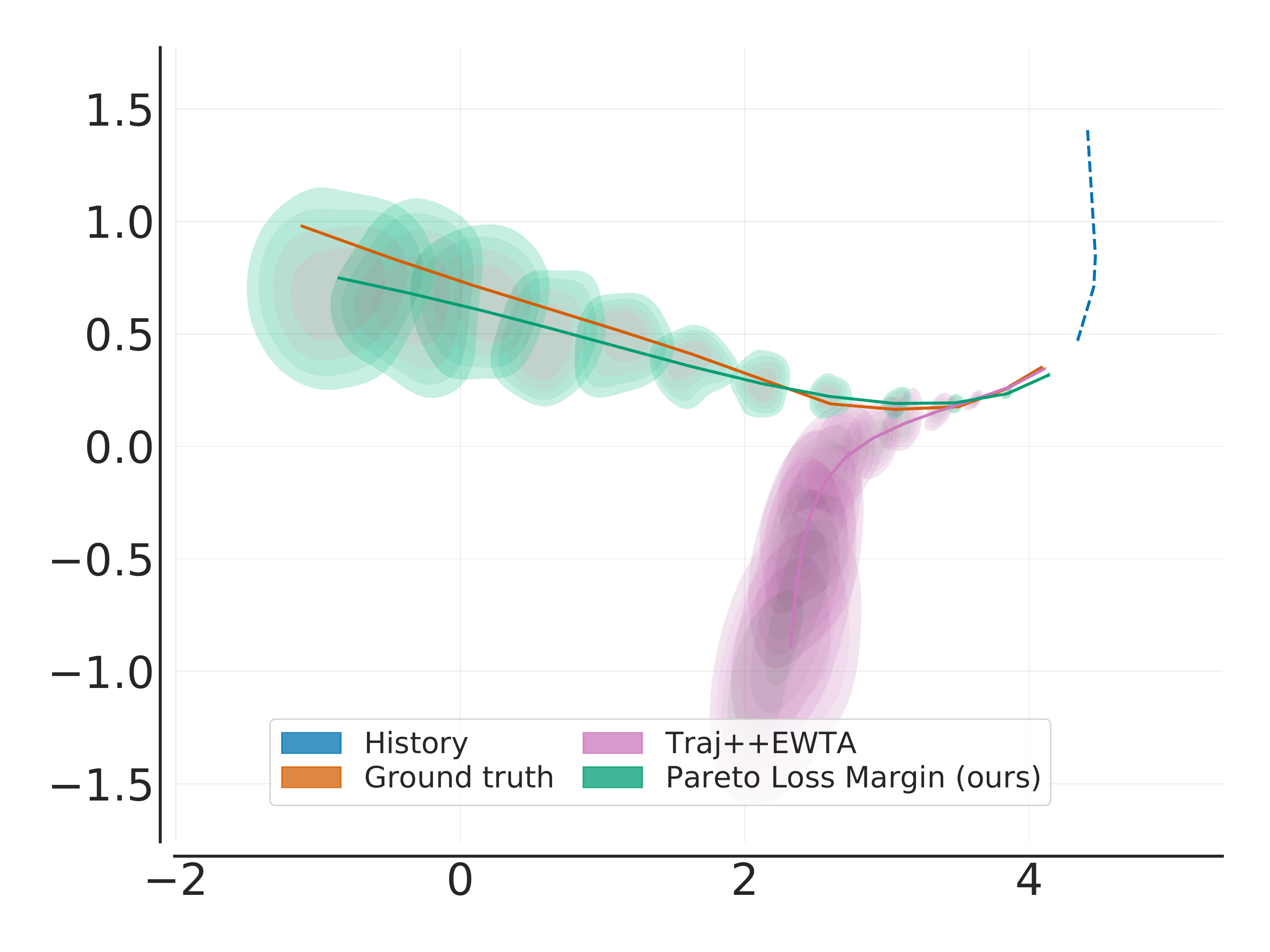}
\includegraphics[width=0.25\textwidth]{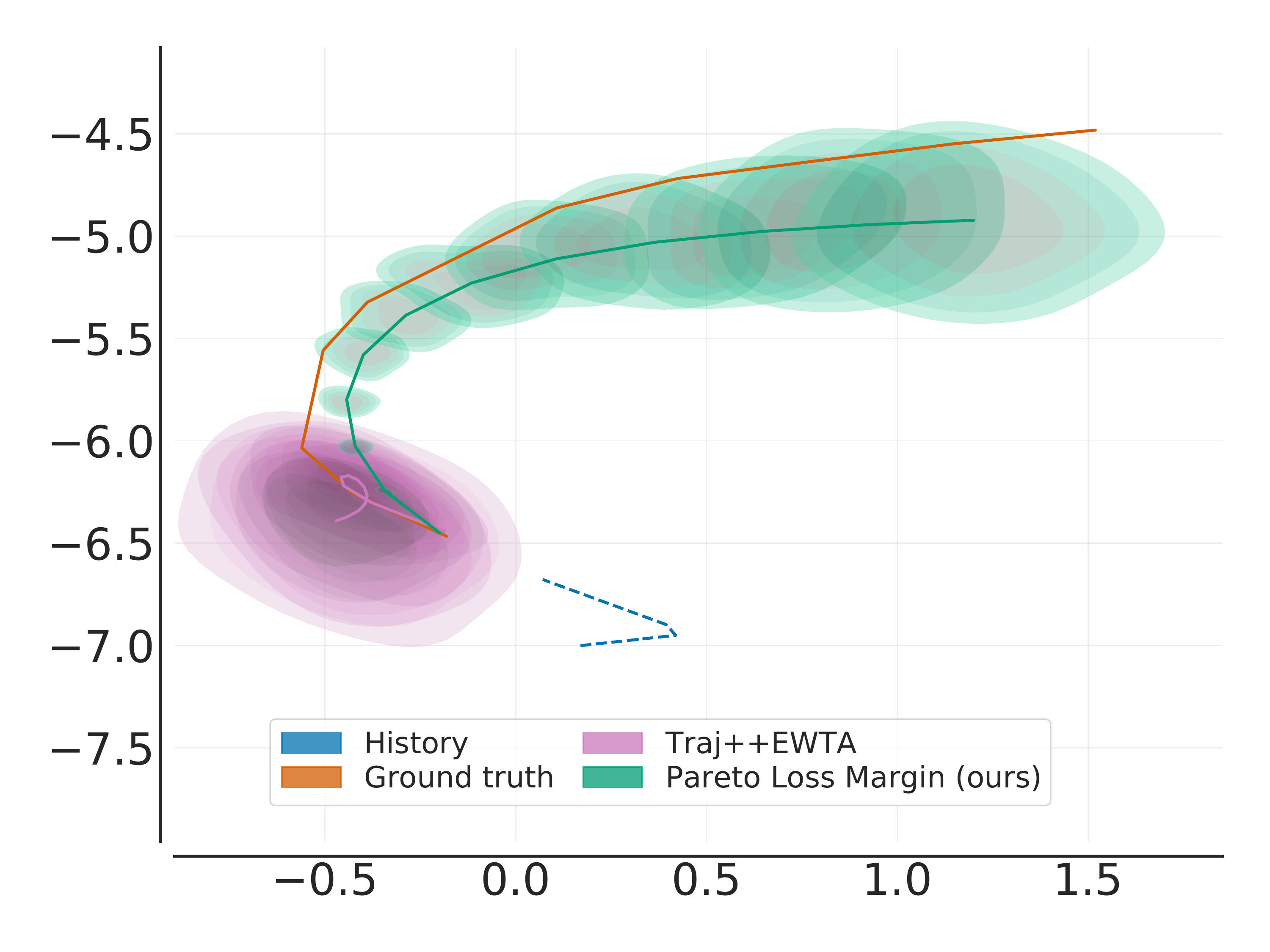}
\includegraphics[width=0.25\textwidth]{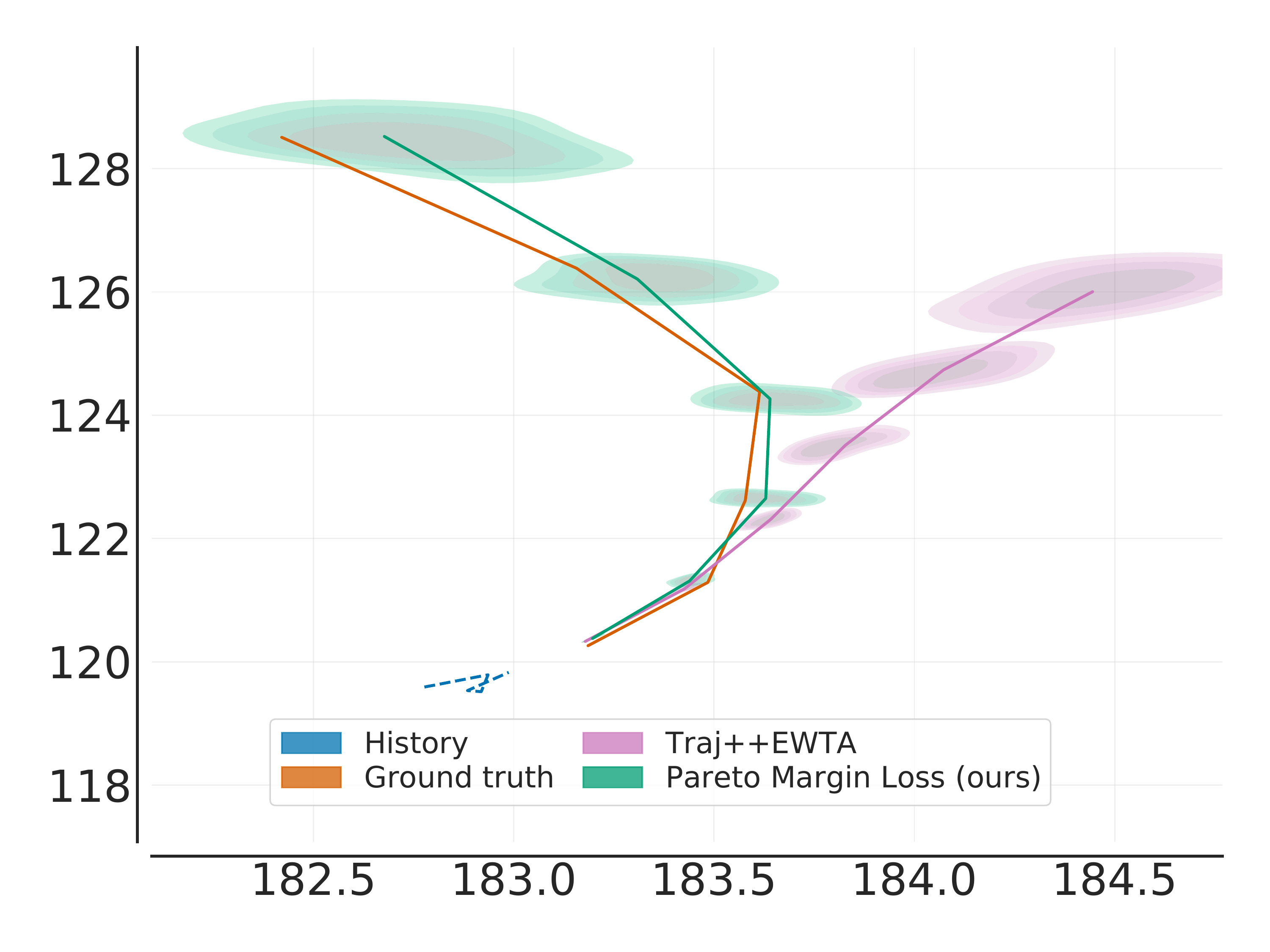}
\includegraphics[width=0.25\textwidth]{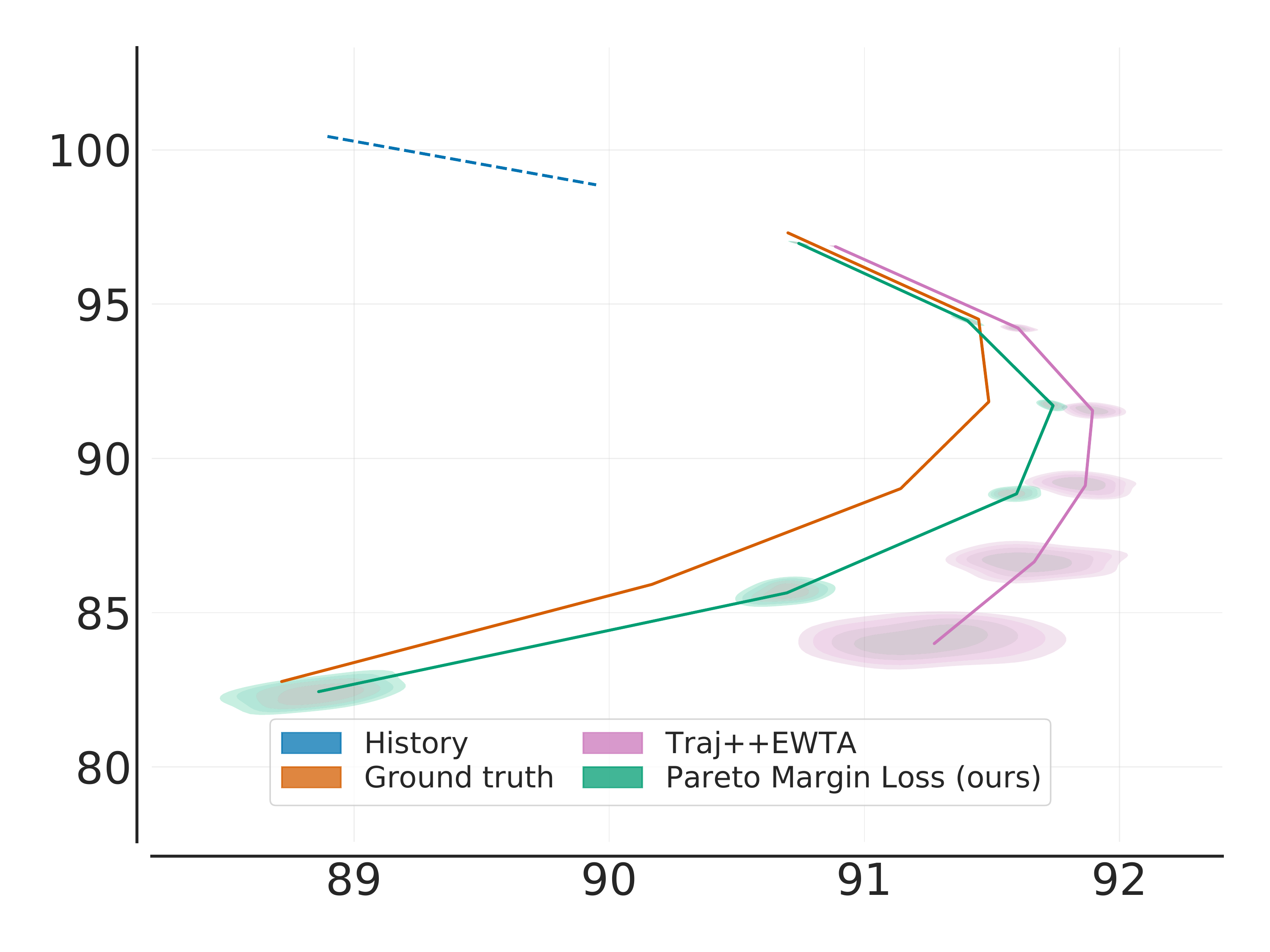}}
\vskip -0.1in
\caption{Visualization of difficult (tail) examples for Electricity (top row left half), Traffic (top row right half), ETH-UCY (bottom row left half) and nuScenes (bottom row right half) datasets. The difficulty in all datasets is captured by a significant departure in behavior with respect to the history. This manifests as sudden increase or decrease in the 1D time series datasets and as high velocity trajectories with sharp turns for the trajectory datasets. These examples represent critical events in real world scenarios where the performance of the model is of utmost importance. Our methods perform significantly better on such examples.}
\label{fig:difficult_examples}
\end{center}
\vskip -0.1in
\end{figure*}

\vspace{-0.05in}
\paragraph{PLM vs Kurtosis loss.}

Kurtosis loss generally performs better on extreme tail metrics, $\mathrm{VaR_{99}}$ and Max. The bi-quadratic behavior of kurtosis puts higher emphasis on far-tail samples. Moreover, the magnitude of kurtosis varies significantly for different distributions, making the choice of hyperparameter (See Eq.\eqref{eq:kbr}) critical. Further analysis on the same is available in Appendix \ref{section:hyperparam_tune}.

PLM is the most consistent method across all tasks improving on both tail and average metrics. As noted by~\cite{mcneil1997estimating} GPD is well suited to model long tail error distributions. PLM rewards examples moving away from the tail towards the mean with significantly lower margin values. PLM margin values saturate beyond a point in the tail providing similar improvements for subsequent tail samples. Visualization of PLM predictions for difficult tail examples can be seen in Fig.~\ref{fig:difficult_examples}.

Kurtosis is sensitive to extreme samples in the tail, while PLM treats most samples in the tail similarly. This manifests in performance as kurtosis loss performing better on $\mathrm{VaR_{99}}$ and Max, and PLM performing better on $\mathrm{VaR_{95}}$ and $\mathrm{VaR_{98}}$. 

This provides guidance on the choice of method as per the objective. Kurtosis Loss can improve the performance in worst case scenarios more significantly. PLM provides less drastic changes to the most extreme values, but it works more effectively throughout the entire distribution. 

\begin{figure}[t!]
\vskip -0.05in
\begin{center}
\centerline{\includegraphics[width=0.6\columnwidth]{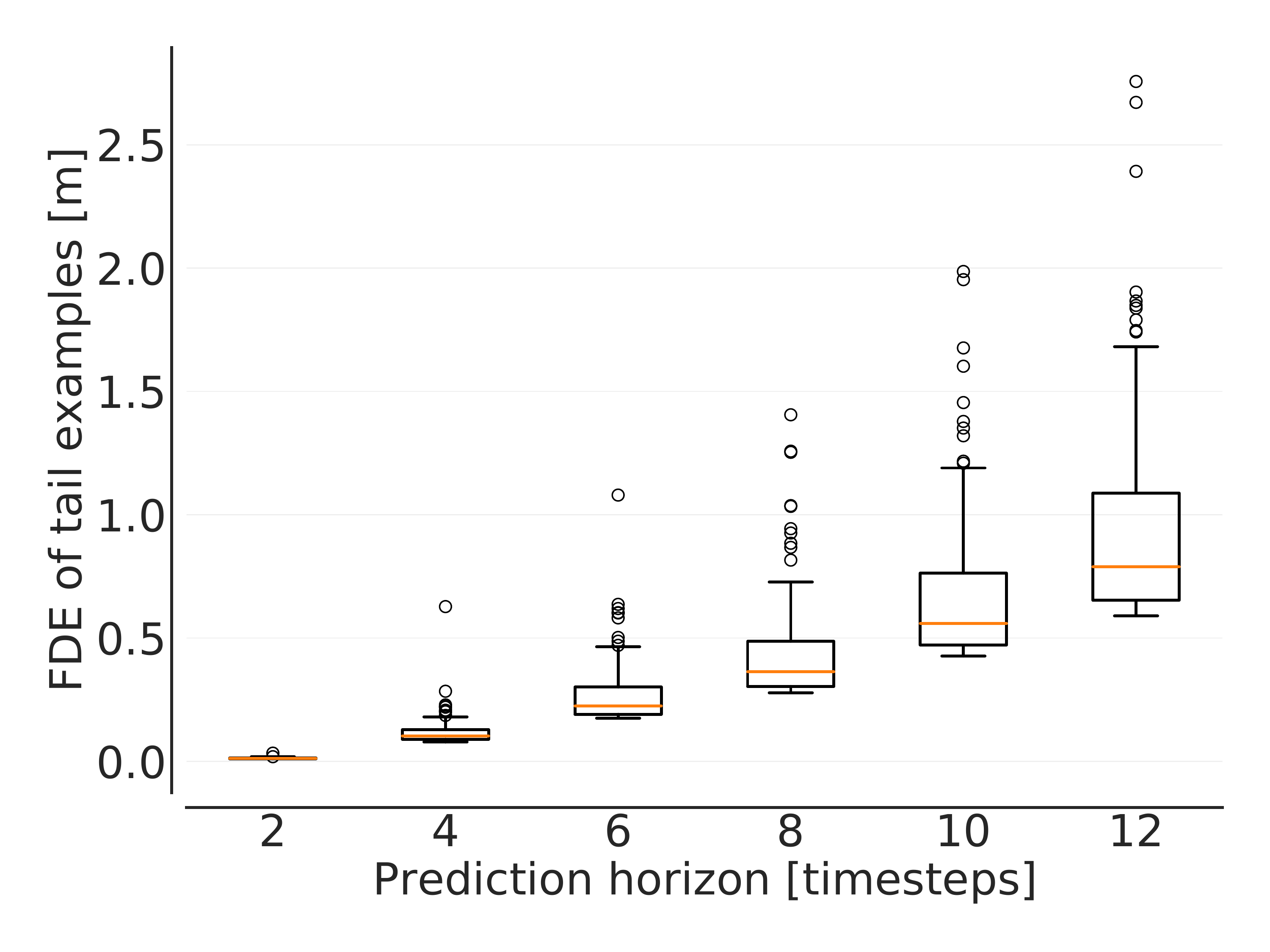}}
\vspace{-0.2in}
\caption{Distribution of the top 5\% most erroneous predictions (FDE) for different prediction horizons for ETH-UCY (Zara1) dataset. Predictions obtained using Trajectron++EWTA. The trend shows that the tail gets much worse as the prediction horizon is increased due to compounded error.}
\label{fig:tail_per_timestep}
\end{center}
\vskip -0.1in
\end{figure}
\vspace{-0.05in}
\paragraph{Tail error and long-term forecasting}
Based on the trajectory forecasting results in Tables~\ref{table:eth_ucy_20guess} and~\ref{table:nuscenes_20guess} we can see that error reduction for tail samples is more visible in FDE than ADE. This indicates that the magnitude of the observed error increases with the prediction horizon. The error accumulates through prediction steps making far-future predictions inherently more difficult. Larger improvements in the FDE indicate that both Kurtosis and Pareto loss ensure that high tail errors (stemming mostly from large, far-future prediction errors measured by FDE) are decreased. 

The inadvertent direction of research in the forecasting domain is aiming at increasing the prediction horizon with high accuracy predictions. As we can see in Fig.~\ref{fig:tail_per_timestep}, the effect of the tail examples is more pronounced with longer prediction horizons. Thus, methods addressing the tail performance will be necessary in order to ensure the practical applicability and reliability of future, long-term prediction.

\vspace{-2mm}
\section{Conclusion}
We address the long-tail problem in deep probabilistic forecasting. We propose Pareto loss (Margin and Weighted) and Kurtosis loss, two novel moment-based loss function approaches increasing emphasis on learning tail examples. We demonstrate their practical effects on two spatiotemporal trajectory datasets and two time series datasets. Our methods achieve significant improvements on tail examples over existing baselines without degrading average performance. Both proposed losses can be integrated with existing approaches in deep probabilistic forecasting to improve their performance on difficult and challenging scenarios. 

Future directions include more principled ways to tune hyperparameters, new approaches to mitigate long tail for long-term forecasting and application to more complex tasks like video prediction. Based on our observations, we suggest evaluating additional tail performance metrics apart from average performance in machine learning task to identify potential long tail issues across different tasks and domains. 

\section*{Acknowledgments}
This work was supported in part by U.S. Department Of Energy, Office of Science, U. S. Army Research Office under Grant W911NF-20-1-0334, Facebook Data Science Award, Google Faculty Award, and NSF Grant \#2037745.
\bibliographystyle{IEEEtran}
\bibliography{main}

\appendix
\newpage
\onecolumn

\section{Dataset description}
\label{section:dataset_description}

The ETH-UCY dataset consists of five subdatasets, each with Bird's-Eye-Views: ETH, Hotel, Univ, Zara1, and Zara2. As is common in the literature~\cite{makansi2021exposing, salzmann2020trajectron++} we present macro-averaged 5-fold cross-validation results in our experiment section. The nuScenes dataset includes 1000 scenes of 20 second length for vehicle trajectories recorded in Boston and Singapore.

The electricity dataset contains electricity consumption data for 370 homes over the period of Jan 1st, 2011 to Dec 31st, 2014 at a sampling interval of 15 minutes. We use the data from Jan 1st, 2011 to Aug 31st, 2011 for training and data from Sep 1st, 2011 to Sep 7th, 2011 for testing. The traffic dataset consists of occupancy values recorded by 963 sensors at a sampling interval of 10 minutes ranging from Jan 1st, 2008 to Mar 30th, 2009. We use data from Jan 1st, 2008 to Jun 15th, 2008 for training and data from Jun 16th, 2008 to Jul 15th, 2008 for testing. Both time series datasets are downsampled to 1 hour for generating examples.

The synthetic datasets are generated as 100 different time series consisting of 960 time steps. Each time series in the Sine dataset is generated using a random offset $\theta$ and a random frequency $\nu$ both selected from a uniform distribution $U[0, 1]$. Then the time series is $sin(2\pi\nu t + \theta)$ where $t$ is the index of the time step. Gaussian and Pareto datasets are generated as order 1 lag autoregressive time series with randomly sampled Gaussian and Pareto noise respectively. Gaussian noise is sampled from a Gaussian distribution with mean 1 and standard deviation 1. Pareto noise is randomly sampled from a Pareto distribution with shape 10 and scaling 1.

\section{Method adaptation}
\label{section:baseline_adapt}

\paragraph{Time Series forecasting}
DeepAR uses Gaussian Negative Log Likelihood as the loss which is unbounded. Due to this many baseline methods need to be adapted in order to be usable. For the same reason, we also need an auxiliary loss ($\hat{l}$). We use MAE loss to fit the GPD, calculate kurtosis, and to calculate the weight terms for Focal and Shrinkage loss. For LDS we treat all labels across time steps as a part of a single distribution. Additionally, to avoid extremely high weights ($\mathcal{O}(10^8)$) in LDS due to the nature of long tail we ensure a minimum probability of $0.001$ for all labels.

\paragraph{Trajectory forecasting}
We adapt Focal Loss and Shrinkage Loss to use EWTA loss~\cite{makansi2019overcoming} in order to be compatible with Trajectron++EWTA base model. LDS was originally proposed for a regression task and we adapt it to the trajectory prediction task in the same way as for the time series task. We use MAE to fit the GPD, due to the Evolving property of EWTA loss.

\section{Implementation details}

\paragraph{Time Series forecasting}
We use the DeepAR implementation from \href{https://github.com/zhykoties/TimeSeries}{https://github.com/zhykoties/TimeSeries} as the base code to run all time series experiments. The original code is an AWS API and not publicly available. The implementation of contrastive loss is taken directly from the source code of \cite{makansi2021exposing}.

\paragraph{Trajectory forecasting}
For all tested base methods in the trajectory forecasting experiments (Trajectron++~\cite{salzmann2020trajectron++} and Trajectron++EWTA~\cite{makansi2021exposing}) we have used the original implementations provided by the original authors of each method. The implementation of contrastive loss is taken directly from the source code of \cite{makansi2021exposing}.

The experiments have been conducted on a machine with 7 RTX 2080 Ti GPUs.

\section{Hyperparameter Tuning}
\label{section:hyperparam_tune}

\begin{figure}
\vskip 0.2in
\begin{center}
\includegraphics[width=0.45\columnwidth]{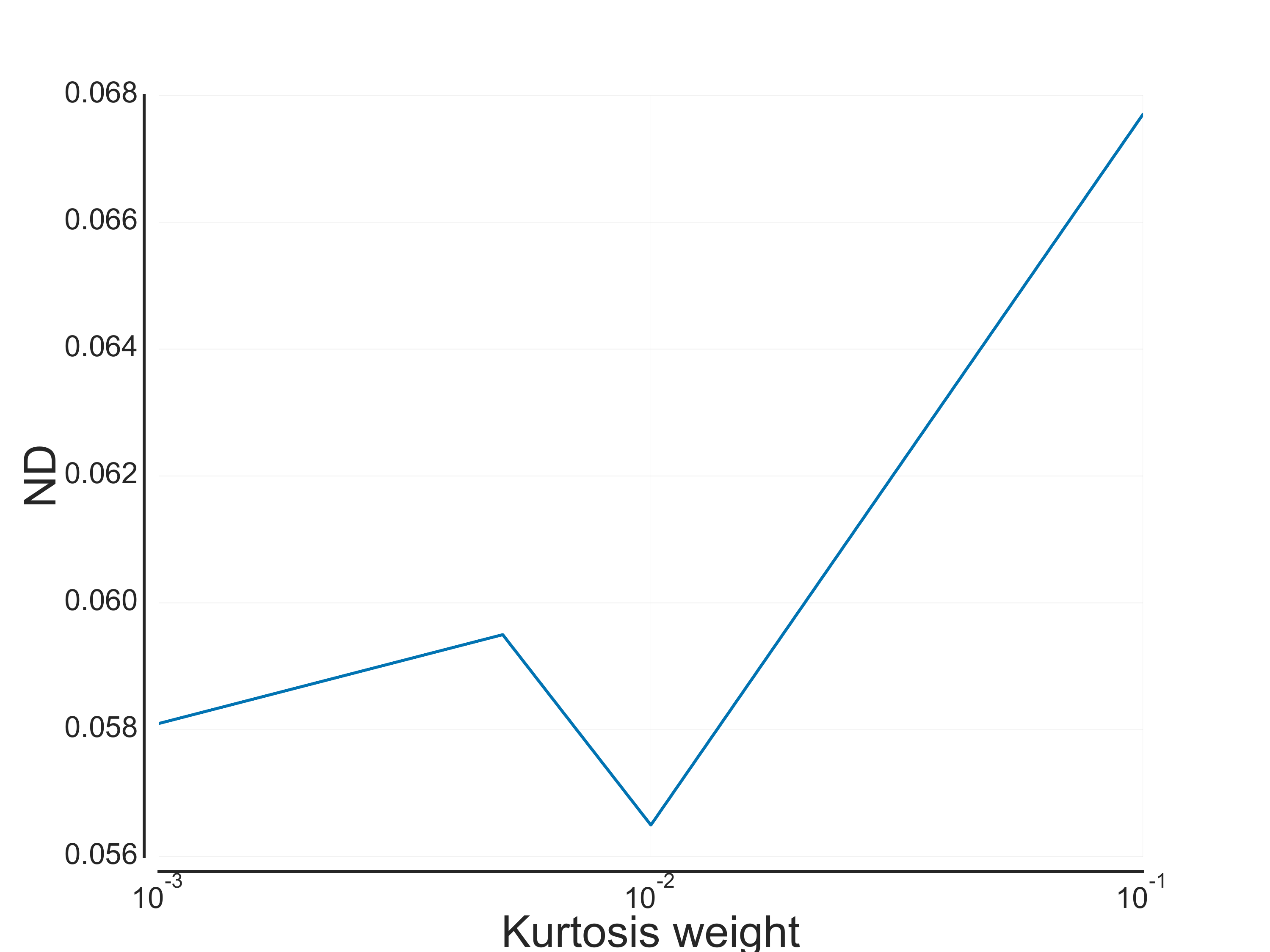}
\includegraphics[width=0.45\columnwidth]{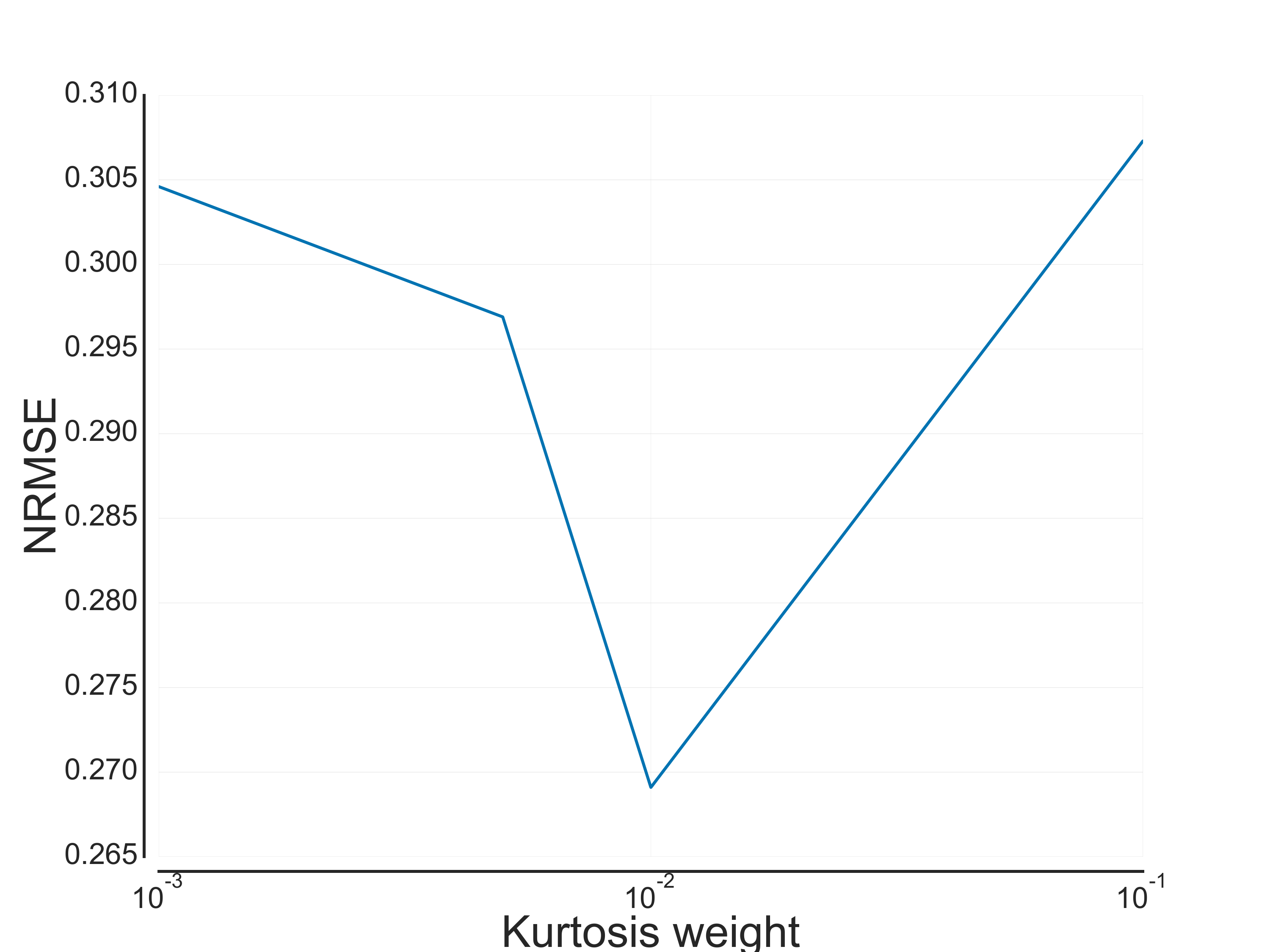}
\caption{Left: Variation of ND by hyperparameter for kurtosis loss. Right: Variation of NRMSE by hyperparameter for kurtosis loss.}
\label{fig:kurt_hyperparam}
\end{center}
\end{figure}

We observe during our experiments that the performance of kurtosis loss is highly dependent on the hyperparameter $\lambda$ (See Eq. \eqref{eq:kbr}). Results for different values of $\lambda$ on the electricity dataset for kurtosis are shown in Table\ref{table:kurt_tuning}. We also show the variation of ND and NRMSE with the hyperparameter value in Figure \ref{fig:kurt_hyperparam}. We can see that there is an optimal value of the hyperparameter and the approach performs worse with higher and lower values. 

For both ETH-UCY and nuScenes datasets we have used $\lambda=0.1$ for Kurtosis loss, and $\lambda=1$ for PLM and PLW. For both electricity and traffic datasets, we use $\lambda=1$ for PLM, $\lambda=0.5$ for PLW and $\lambda=0.01$ for Kurtosis loss.

\begin{table*}[h]
\caption{Electricity Dataset Evaluation using DeepAR (ND/NRMSE) and different Kurtosis loss hyperparameters. The value of $\lambda$ is denoted in [] with the method name}
\label{table:kurt_tuning}
\begin{center}
\begin{small}
\begin{sc}
\begin{tabular}{lccccccccc}
\toprule
Method & Metric & Mean$\downarrow$ & $VAR_{95}\downarrow$ & $VAR_{98}\downarrow$ & $VAR_{99}\downarrow$ & Max$\downarrow$ & Kurtosis$\downarrow$ & Skew$\downarrow$ \\
\midrule
DeepAR & ND & 0.0584 & 0.0796 & 0.2312 & 0.4429 & 4.1520 & 426.5906 & 18.4057\\
& NRMSE & 0.2953 & 0.0972 & 0.2595 & 0.5263 & 5.4950 & 470.8968 & 19.4827\\
\midrule
+ Kurtosis Loss [0.001] & ND & 0.0581 & 0.0815 & 0.2087 & 0.3936 & 4.2381 & 488.7306 & 19.8207\\
& NRMSE & 0.3046 & 0.1014 & 0.2325 & 0.4756 & 5.7144 & 529.7499 & 20.7713\\
\midrule
+ Kurtosis Loss [0.005] & ND & 0.0574 & 0.0767 & 0.2147 & 0.4138 & 3.6767 & 351.3378 & 16.7597\\
& NRMSE & 0.2843 & 0.0999 & 0.2617 & 0.4792 & 5.0062 & 417.0575 & 18.3039\\
\midrule
+ Kurtosis Loss [0.01] & ND & 0.0567 & 0.0842 & 0.2151 & 0.4120 & 3.2738 & 300.3517 & 15.4597\\
& NRMSE & 0.2631 & 0.1046 & 0.2732 & 0.4779 & 4.2613 & 339.3773 & 16.4892\\
\midrule
+ Kurtosis Loss [0.1] & ND & 0.0677 & 0.0954 & 0.2269 & 0.4579 & 3.8772 & 312.7331 & 16.0062\\
& NRMSE & 0.3073 & 0.1184 & 0.2768 & 0.5419 & 5.1345 & 334.8358 & 16.3366\\
\bottomrule
\end{tabular}
\end{sc}
\end{small}
\end{center}
\vskip -0.1in
\end{table*}

\section{Long tail severity}\label{section:tail_severity}
In Table~\ref{table:tail_severity} we present the numerical values representing the approximate long-tailedness for each of the datasets. Larger value indicates a longer tail.

\begin{table}[H]
\vskip 0in
\caption{An approximate long-tailedness (based on Eq.~\eqref{eq:tail_severity}) in the performance of base model on different datasets. Higher number indicates a longer tail. Trajectory datasets have shorter tail than 1D timeseries datasets.}
\label{table:tail_severity}
\vskip 0.15in
\begin{center}
\begin{small}
\begin{sc}
\begin{tabular}{lcc}
\toprule
Dataset & Metric & Long-tailedness\\
\midrule
Electricity & ND & 15.56\\
Traffic & ND & 45.08\\
ETH-UCY & FDE & 7.96\\
nuScenes & FDE & 10.81\\
\bottomrule
\end{tabular}
\end{sc}
\end{small}
\end{center}
\vskip 0in
\end{table}

\section{Pareto and Kurtosis}\label{section:pareto_kurtosis}
\begin{figure}[h]
\vskip 0.2in
\begin{center}
\includegraphics[width=0.45\columnwidth]{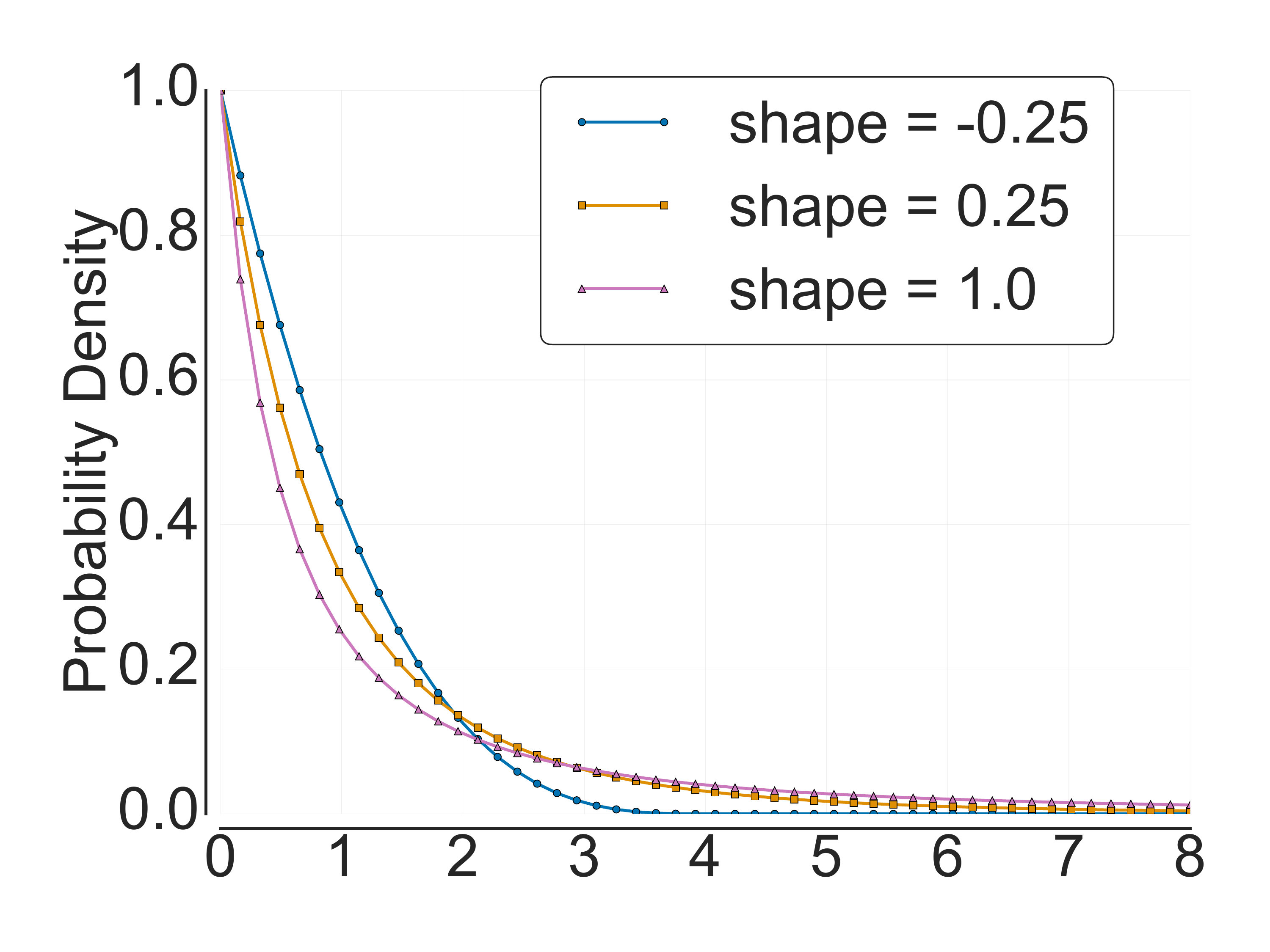}
\includegraphics[width=0.45\columnwidth]{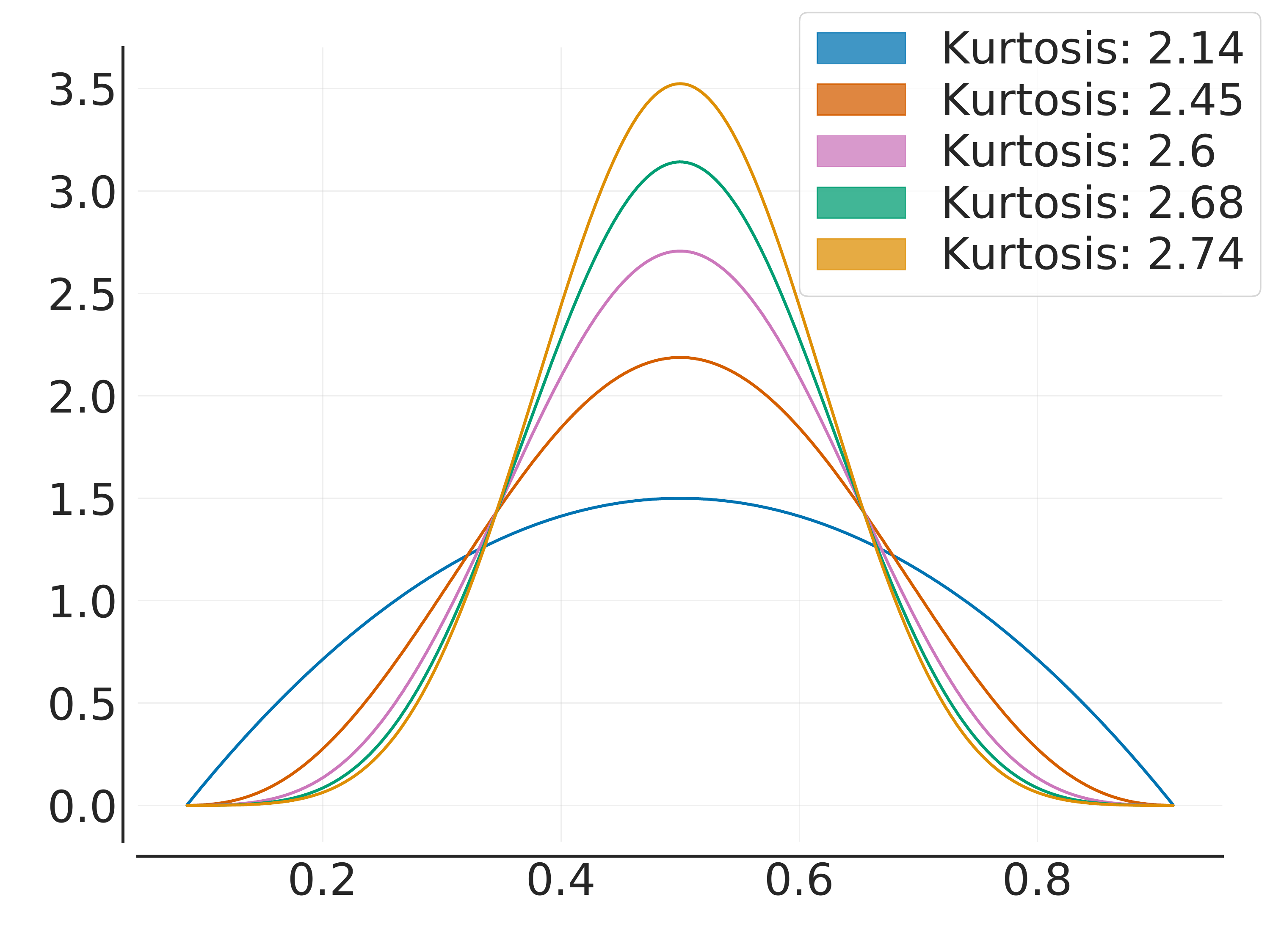}
\caption{Left: Generalized Pareto distributions with different shape parameters ($\eta=1$). Right: Illustrating the variation of kurtosis on distributions with the same mean.}
\label{fig:gpd_shape}
\end{center}
\end{figure}

Figure~\ref{fig:gpd_shape} illustrates different GPDs for different shape parameter values. Higher shape value models more severe tail behavior.

\section{Long tail error distribution}\label{section:error_distribution}

\begin{figure}[h!]
\vskip -0.1in
\begin{center}
\centerline{
\includegraphics[width=0.45\columnwidth]{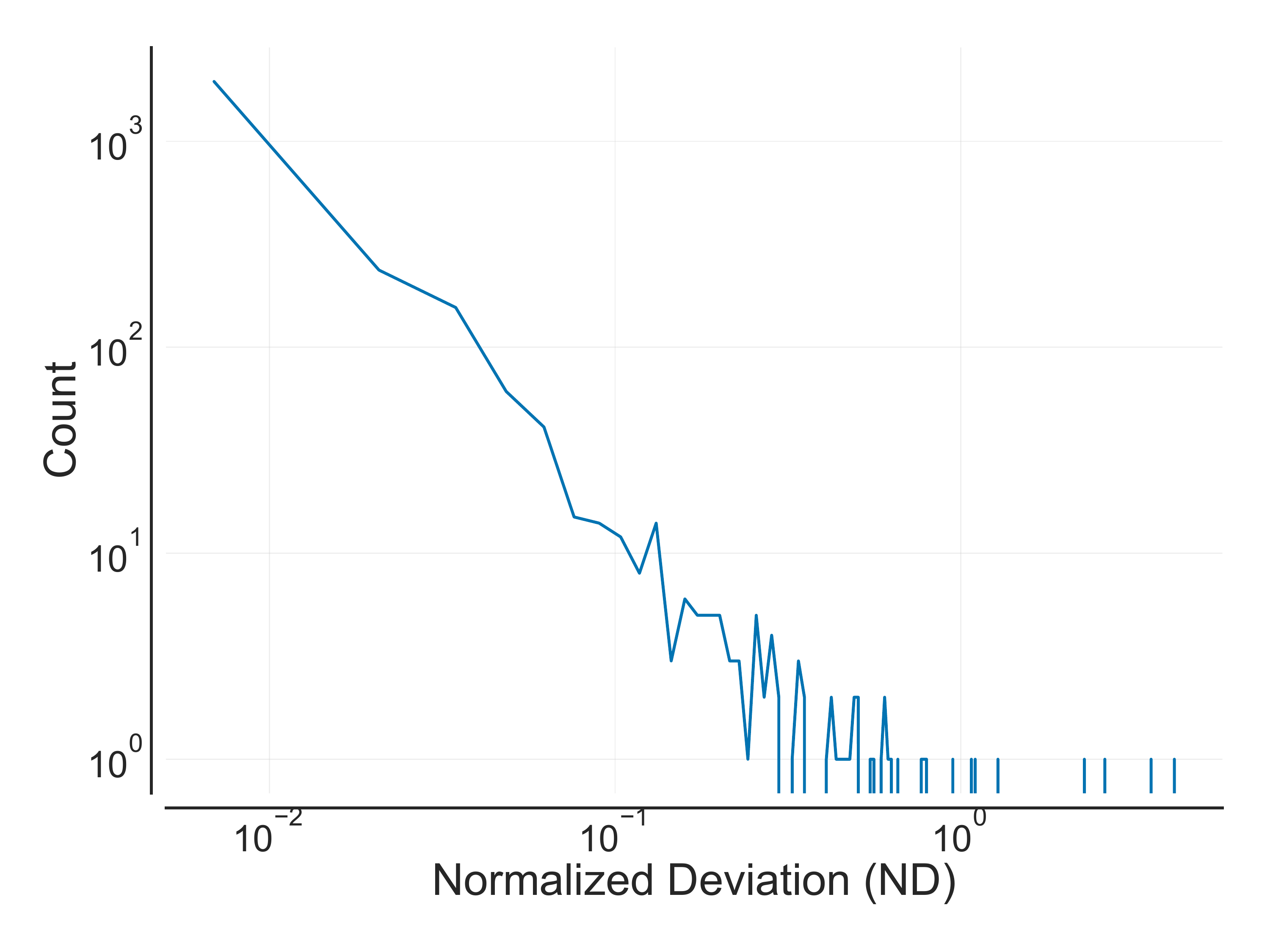}
\includegraphics[width=0.45\columnwidth]{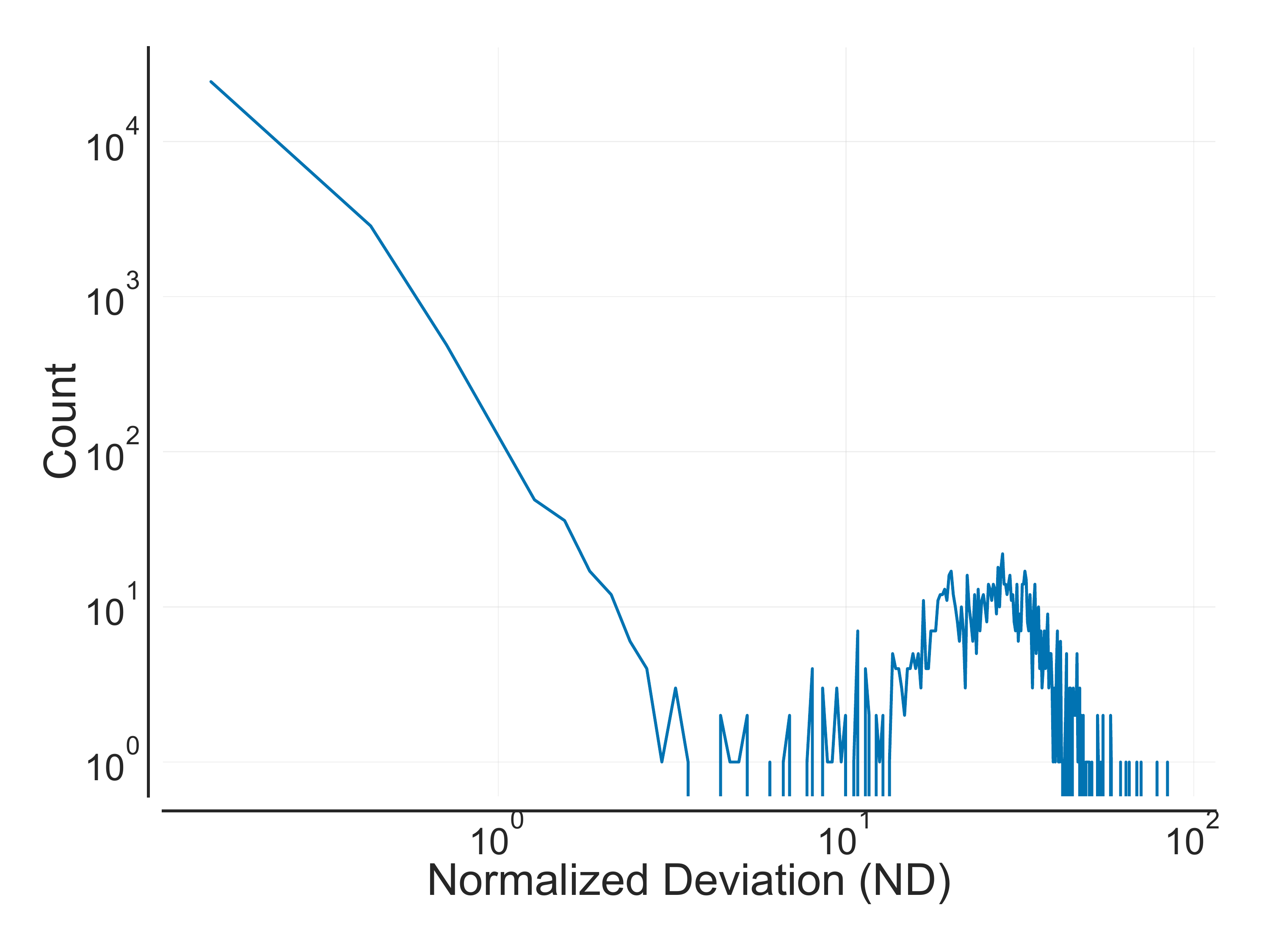}
}
\centerline{
\includegraphics[width=0.45\columnwidth]{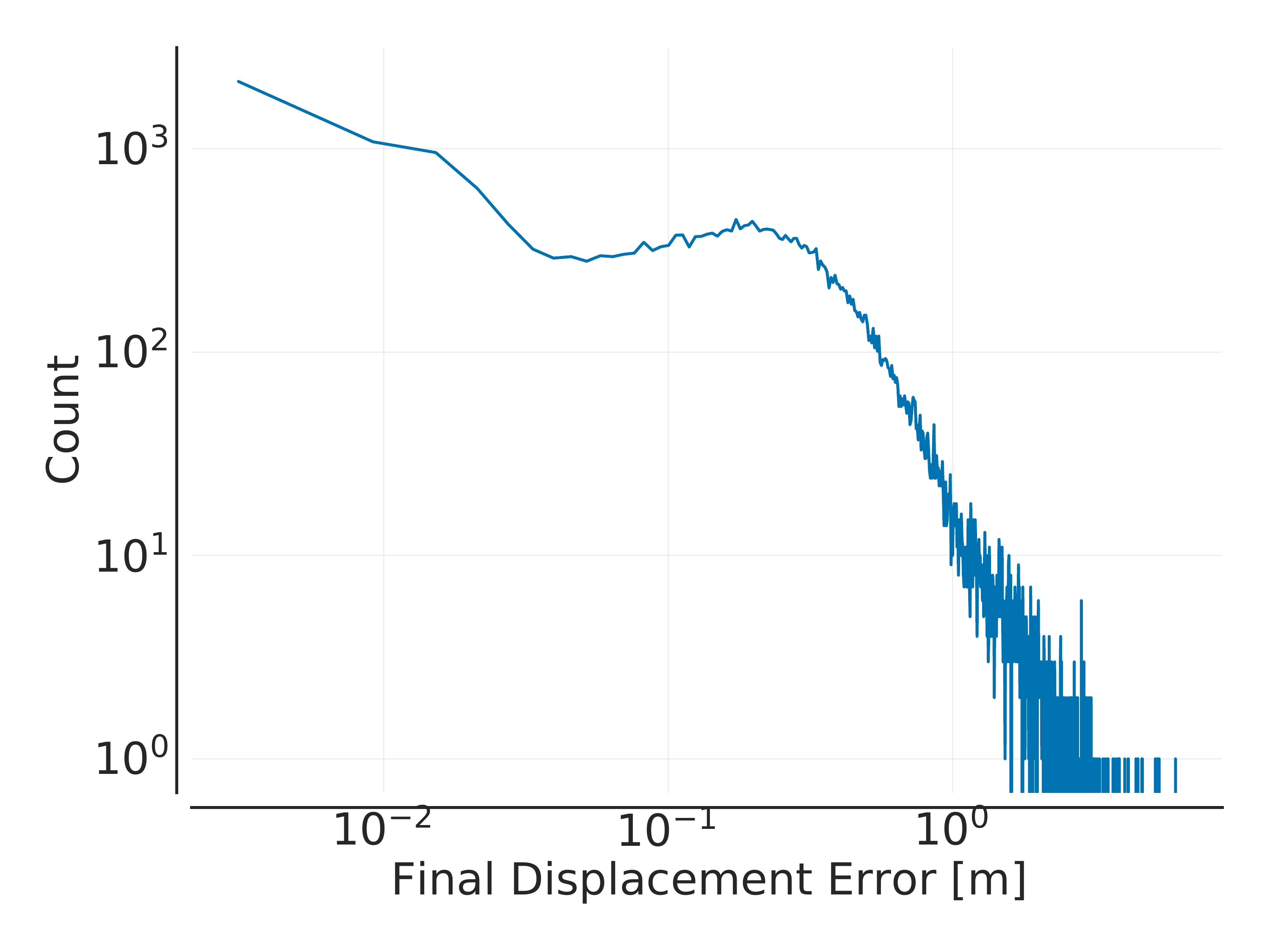}
\includegraphics[width=0.45\columnwidth]{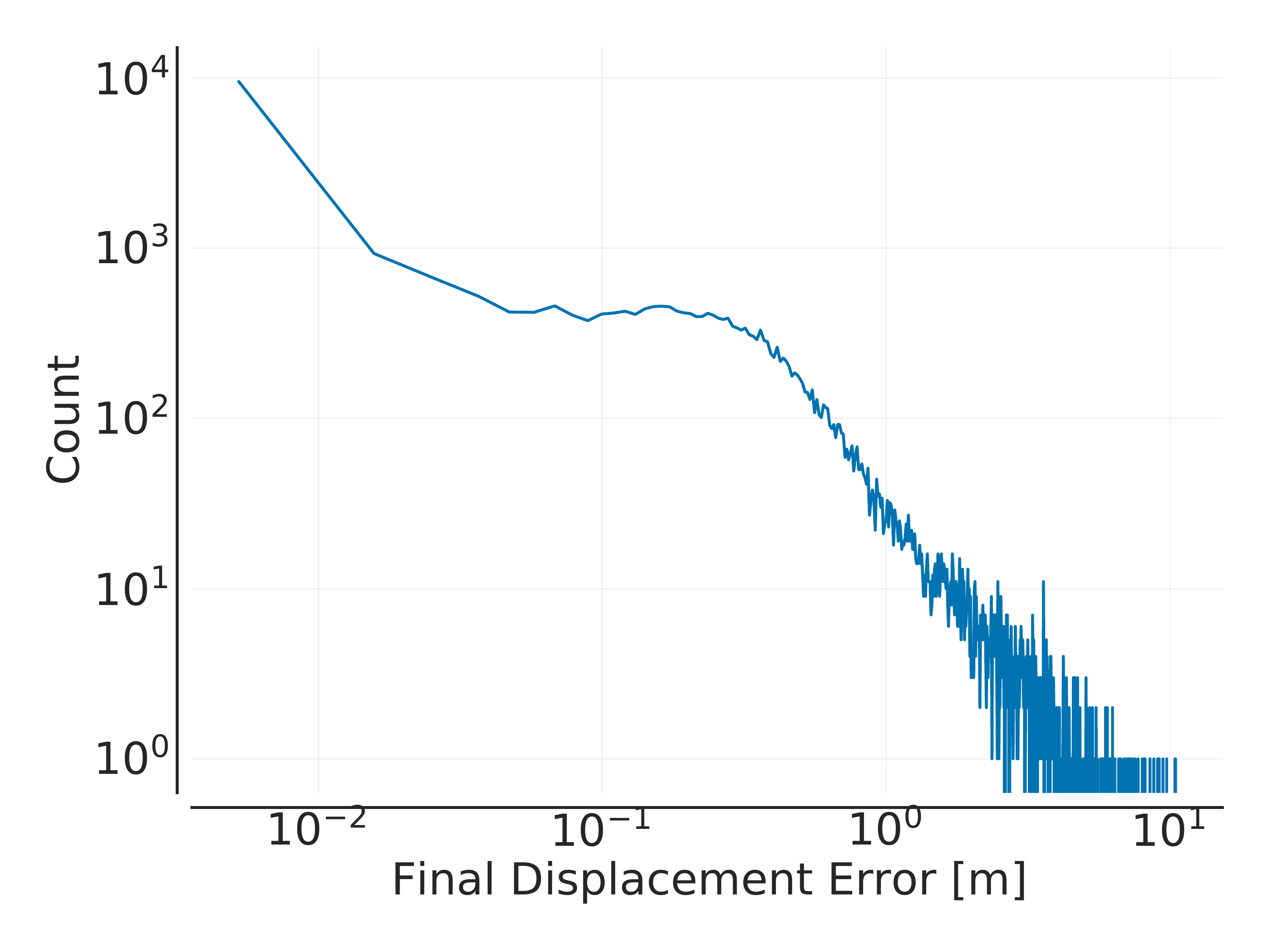}
}
\vskip -0.1in
\caption{Log log plots of error distribution of base model for the time series [Electricity: Top left, Traffic: Top right] and trajectory [ETH-UCY: Bottom left, nuScenes: Bottom right] forecasting datasets.} 
\label{fig:log_log_error}
\end{center}
\vskip -0.1in
\end{figure}

In Fig.~\ref{fig:log_log_error} we can see log-log plots of the error distributions of base model for each of the datasets. We can see each distribution exhibits a long tail behavior.

\section{Synthetic datasets}
\label{section:syn_dataset_analysis}

We present complete results of our experiments on the synthetic datasets in Table \ref{table:deepar_synthetic}. We ran our methods, kurtosis loss, and PLM on these datasets as well. Both our methods show significant tail improvements over the base model across all datasets.

\begin{table*}[h]
\caption{Performance on the Synthetic Datasets (ND/NRMSE).}
\label{table:deepar_synthetic}
\vskip 0.15in
\begin{center}
\begin{small}
\begin{sc}
\begin{tabular}{lcccccccc}
\toprule
Method & Metric & Mean$\downarrow$ & $VAR_{95}\downarrow$ & $VAR_{98}\downarrow$ & $VAR_{99}\downarrow$ & Max$\downarrow$ & Kurtosis$\downarrow$ & Skew$\downarrow$ \\
\midrule
 \multicolumn{9}{c}{Sine Dataset}\\
 \midrule
AutoReg & ND & 1.2255 & 2.162 & 2.7088 & 2.9306 & 3.1271 & -0.1565 & 0.1905\\
& NRMSE & 1.5078 & 2.3134 & 2.7204 & 2.9379 & 3.1271 & -0.56 & -0.0369\\
\midrule
DeepAR & ND & 0.0513 & 0.1721 & 0.316 & 0.5913 & 1.5744 & 71.9164 & 7.9019\\
& NRMSE & 0.1534 & 0.2009 & 0.3507 & 0.6199 & 1.654 & 64.4497 & 7.4304\\
\midrule
+ Kurtosis Loss & ND & 0.0455 & 0.1412 & 0.2914 & 0.447 & 1.5571 & 90.6313 & 8.6956\\
& NRMSE & 0.133 & 0.1624 & 0.3455 & 0.5387 & 1.5571 & 76.7183 & 7.9383\\
\midrule
+ Pareto Loss & ND & 0.0462 & 0.1326 & 0.3014 & 0.7151 & 1.582 & 78.6768 & 8.4086\\
Margin & NRMSE & 0.1517 & 0.1563 & 0.3551 & 0.737 & 1.7522 & 72.0235 & 7.9663\\
\midrule
 \multicolumn{9}{c}{Gaussian Dataset}\\
\midrule
AutoReg & ND & 0.573 & 1.0225 & 1.3334 & 1.6226 & 27.6956 & 845.0732 & 26.4337\\
& NRMSE & 1.2705 & 1.1212 & 1.4045 & 1.6815 & 39.7474 & 1010.198 & 29.748\\
\midrule
DeepAR & ND & 0.4379 & 0.705 & 0.7908 & 0.8651 & 1.1362 & 0.8225 & 0.7469\\
& NRMSE & 0.5518 & 0.8172 & 0.9246 & 0.9908 & 1.3009 & 0.5562 & 0.65\\
\midrule
+ Kurtosis Loss & ND & 0.4378 & 0.704 & 0.7973 & 0.8597 & 1.1294 & 0.8037 & 0.7418\\
& NRMSE & 0.5518 & 0.8191 & 0.9255 & 0.9865 & 1.2951 & 0.539 & 0.6449\\
\midrule
+ Pareto Loss & ND & 0.4391 & 0.7023 & 0.7946 & 0.8674 & 1.1069 & 0.7813 & 0.7352\\
Margin & NRMSE & 0.5534 & 0.8194 & 0.9232 & 0.9889 & 1.2786 & 0.4985 & 0.6333\\
\midrule
 \multicolumn{9}{c}{Pareto Dataset}\\
\midrule
AutoReg & ND & 1.9377 & 1.1748 & 1.7039 & 2.4782 & 2113.7503 & 2116.5018 & 44.2477\\
& NRMSE & 81.1652 & 1.4027 & 1.9856 & 2.7312 & 4069.3972 & 2204.8078 & 45.3437\\
\midrule
DeepAR & ND & 0.4416 & 0.8336 & 1.0317 & 1.1763 & 2.015 & 6.9242 & 2.036\\
& NRMSE & 0.6349 & 1.1511 & 1.4295 & 1.6688 & 2.8327 & 7.0681 & 2.1547\\
\midrule
+ Kurtosis Loss & ND & 0.4413 & 0.8345 & 1.0295 & 1.1738 & 2.0326 & 6.8831 & 2.0318\\
& NRMSE & 0.6352 & 1.1541 & 1.4305 & 1.6653 & 2.8335 & 6.9941 & 2.144\\
\midrule
+ Pareto Loss & ND & 0.4394 & 0.8497 & 1.0473 & 1.1955 & 2.086 & 6.6526 & 2.0185\\
Margin & NRMSE & 0.6397 & 1.1694 & 1.447 & 1.6735 & 2.845 & 6.4693 & 2.0711\\
\bottomrule
\end{tabular}
\end{sc}
\end{small}
\end{center}
\vskip 0in
\end{table*}

\end{document}